\documentclass[journal]{IEEEtran}
\usepackage{times}
\usepackage{epsfig}
\usepackage{graphicx}
\usepackage{amsmath}
\usepackage{amssymb}
\usepackage{colortbl}
\usepackage{booktabs} % For formal tables

\usepackage{lipsum}

\newcommand\blfootnote[1]{%
  \begingroup
  \renewcommand\thefootnote{}\footnote{#1}%
  \addtocounter{footnote}{-1}%
  \endgroup
}

\usepackage[breaklinks=true,bookmarks=false]{hyperref}

\begin{document}
\title{One Pixel Attack for Fooling \\Deep Neural Networks}

\author{Jiawei~Su*,
        Danilo~Vasconcellos~Vargas*
        and~Kouichi~Sakurai% <-this % stops a space
\thanks{Authors are with the Graduate School/Faculty of Information Science and Electrical Engineering, Kyushu University, Japan. The third author is also affiliated to Advanced Telecommunications Research Institute International (ATR).}
\thanks{The official version of this article has been published in IEEE Transactions on Evolutionary Computation \cite{1989}, which can be accessed through the following link: https://ieeexplore.ieee.org/abstract/document/8601309} 
}% <-this % stops a space

\maketitle

\begin{abstract}
   Recent research has revealed that the output of Deep Neural Networks (DNN) can be easily altered by adding relatively small perturbations to the input vector. In this paper, we analyze an attack in an extremely limited scenario where only one pixel can be modified. For that we propose a novel method for generating one-pixel adversarial perturbations based on differential evolution (DE). It requires less adversarial information (a black-box attack) and can fool more types of networks due to the inherent features of DE. The results show that 67.97$\%$ of the natural images in Kaggle CIFAR-10 test dataset and 16.04$\%$ of the ImageNet (ILSVRC 2012) test images can be perturbed to at least one target class by modifying just one pixel with 74.03$\%$ and 22.91$\%$ confidence on average. We also show the same vulnerability on the original CIFAR-10 dataset. Thus, the proposed attack explores a different take on adversarial machine learning in an extreme limited scenario, showing that current DNNs are also vulnerable to such low dimension attacks. Besides, we also illustrate an important application of DE (or broadly speaking, evolutionary computation) in the domain of adversarial machine learning: creating tools that can effectively generate low-cost adversarial attacks against neural networks for evaluating robustness.
 \blfootnote{*Both authors have equal contribution.}
\end{abstract}


% \fi
\IEEEpeerreviewmaketitle

\section{Introduction}

\begin{figure}[t]
\begin{center}
\includegraphics[width=0.8\linewidth]{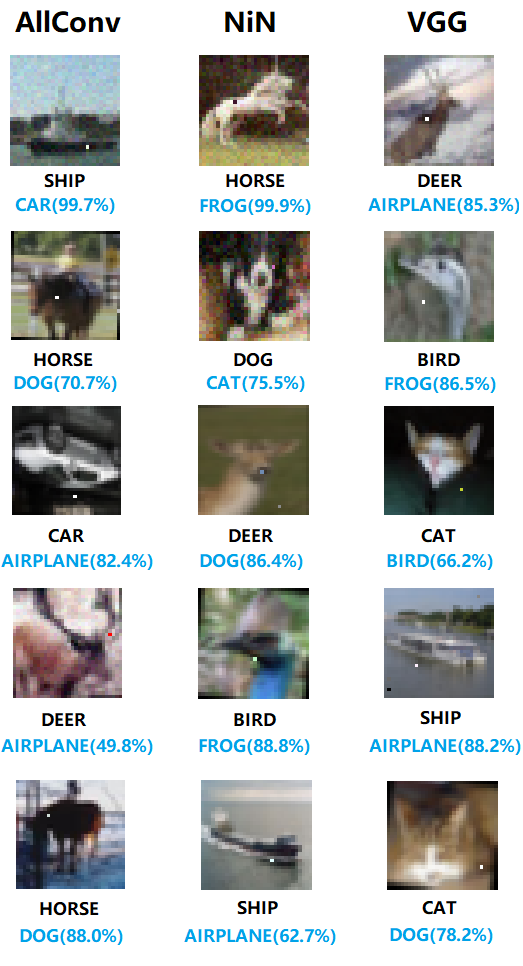}
\end{center}
   \caption{One-pixel attacks created with the proposed algorithm that successfully fooled three types of DNNs trained on CIFAR-10 dataset: The All convolutional network (AllConv), Network in network (NiN) and VGG. The original class labels are in black color while the target class labels and the corresponding confidence are given below.}
\label{our_results}
\label{fig:long}
\label{fig:onecol} 
\end{figure}

\begin{figure}[t]
\begin{center}
\includegraphics[width=0.8\linewidth]{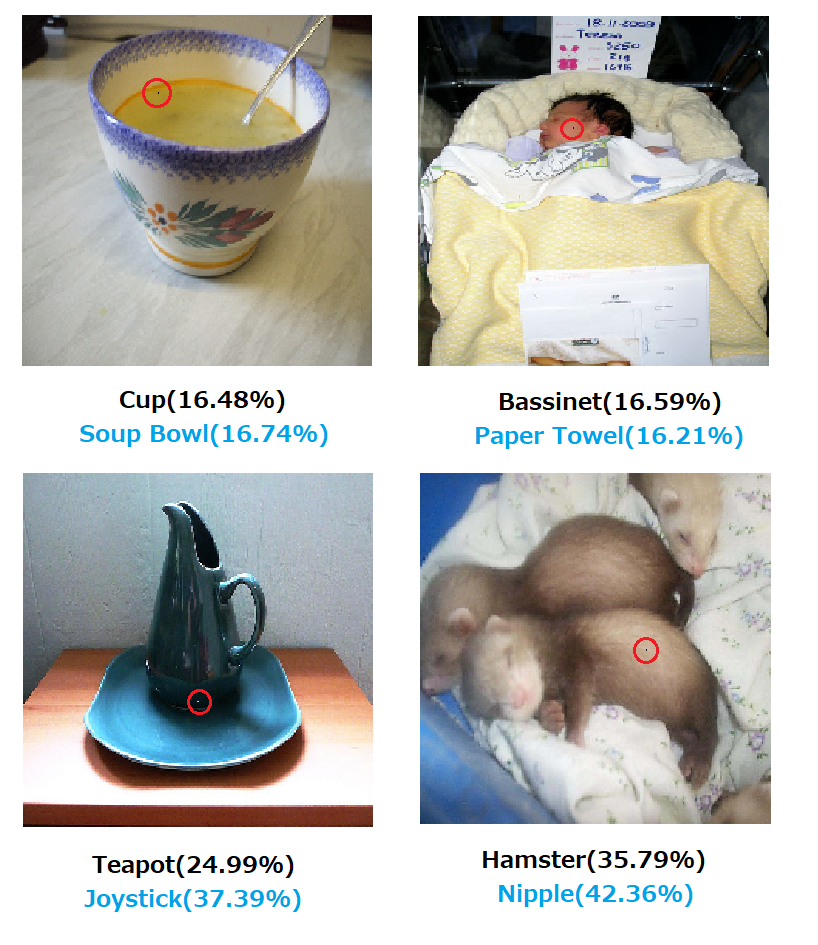}
\end{center}
   \caption{One-pixel attacks on ImageNet dataset where the modified pixels are highlighted with red circles. The original class labels are in black color while the target class labels and their corresponding confidence are given below. }
\label{our_results}
\label{fig:long}
\label{fig:onecol} 
\end{figure}

\begin{figure}[t]
\begin{center}
\includegraphics[width=0.7\linewidth]{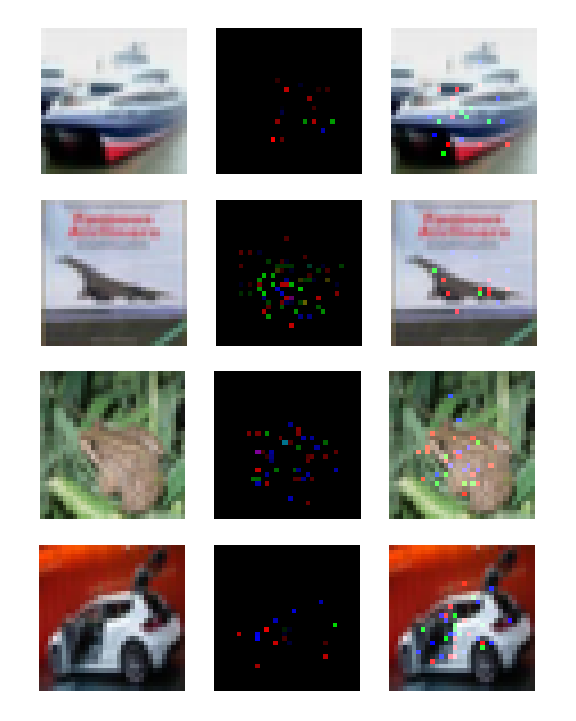}
\end{center}
   \caption{An illustration of the adversarial images generated by using Jacobian saliency-map approach \cite {1}. The perturbation is conducted on about 4$\%$ of the total pixels and can be obvious to human eyes. Since the adversarial pixel perturbation has become a common way of generating adversarial images, such abnormal ``noise'' might be recognized with expertise.}
\label{fig:long}
\label{fig:aaa}
\label{fig:onecol} 
\end{figure}

\IEEEPARstart{I}{n} the domain of image recognition, DNN-based approach has outperform traditional image processing techniques, achieving even human-competitive results \cite {9}. 
However, several studies have revealed that artificial perturbations on natural images can easily make DNN misclassify and accordingly proposed effective algorithms for generating such samples called ``adversarial images'' \cite {4}\cite {2}\cite {1}\cite {3}. 
A common idea for creating adversarial images is adding a tiny amount of well-tuned additive perturbation, which is expected to be imperceptible to human eyes, to a correctly classified natural image. 
Such modification can cause the classifier to label the modified image as a completely different class. 
Unfortunately, most of the previous attacks did not consider extremely limited scenarios for adversarial attacks, namely the modifications might be excessive (i.e., the amount of modified pixels is fairly large) such that it may be perceptible to human eyes (see Figure~\ref{fig:aaa} for an example). 
Additionally, investigating adversarial images created under extremely limited scenarios might give new insights about the geometrical characteristics and overall behavior of DNN's model in high dimensional space \cite{27}. 
For example, the characteristics of adversarial images close to the decision boundaries can help describing the boundaries' shape.

In this paper, by perturbing only one pixel with differential evolution, we propose a black-box DNN attack in a scenario where the only information available is the probability labels (Figure~1 and \ref{our_results})
Our proposal has mainly the following advantages compared to previous works:
 
\begin{itemize}

\item \textbf{Effectiveness} - On Kaggle CIFAR-10 dataset, being able to launch non-targeted attacks by only modifying one pixel on three common deep neural network structures with $68.71\%$, $71.66\%$ and $63.53\%$ success rates. We additionally find that each natural image can be perturbed to $1.8$, $2.1$ and $1.5$ other classes.
On the original CIFAR-10 dataset with a more limited attack scenario, we show $22.60\%$, $35.20\%$ and $31.40\%$ success rates.
On ImageNet dataset, non-targeted attacking the BVLC AlexNet model also by changing one pixel shows that $16.04\%$ of the test images can be attacked.
%Being able to launch non-targeted attacks by only modifying one pixel with $70.97\%$ success rate and $97.47\%$ probability label of target classes on average. 
%\textbf{The effectiveness of conducting non-target attack using few-pixel attack.} We launch the proposed few-pixel attack to three common deep neural network structures and show that with only 1 pixel modification, there are 73.80$\%$, 73.04$\%$ and 66.08$\%$ of the test images that can be perturbed to at least one target class on three networks. 
%We call such images as sensitive images and find that the non-sensitive images are generally even much rarer than sensitive images even if limiting the strength of perturbation to such a small scope.
%\textbf{The number of target classes that a natural image can camouflage.} For three types of networks, we find that each natural image can be perturbed to 2.3, 2.5 and 1.9 other classes. In particular, there are $25.9\%$ , $22.1\%$ and $19.9\%$ of the images that can be perturbed to 1, 2, 3 target classes on average. 
\item \textbf{Semi-Black-Box Attack} - Requires only black-box feedback (probability labels) but no inner information of target DNNs such as gradients and network structures. Our method is also simpler  than existing approaches since it does not abstract the problem of searching perturbation to any explicit target functions but directly focus on increasing the probability label values of the target classes.
\item \textbf{Flexibility} - Can attack more types of DNNs (e.g., networks that are not differentiable or when the gradient calculation is difficult).
\end{itemize}

Regarding the extremely limited one-pixel attack scenario, there are several main reasons why we consider it: 
\begin{itemize}

\item \textbf{Analyze the Vicinity of Natural Images} - Geometrically, several previous works have analyzed the vicinity of natural images by limiting the length of perturbation vector. For example, the universal perturbation adds small value to each pixel such that it searches the adversarial images in a sphere region around the natural image \cite{28}. On the other side, the proposed few-pixel perturbations can be regarded as cutting the input space using very low-dimensional slices, which is a different way of exploring the features of high dimensional DNN input space. Among them, one-pixel attack is an extreme case of several-pixel attack. Theoretically, it can give geometrical insight to the understanding of CNN input space, in contrast to another extreme case: universal adversarial perturbation \cite{28} that modifies every pixel.

\item \textbf{A Measure of Perceptiveness} - The attack can be effective for hiding adversarial modification in practice. To the best of our knowledge, none of the previous works can guarantee that the perturbation made can be completely imperceptible. A direct way of mitigating this problem is to limit the amount of modifications to as few as possible. 
Specifically, instead of theoretically proposing additional constraints or considering more complex cost functions for conducting perturbation, we propose an empirical solution by limiting the number of pixels that can be modified. In other words, we use the number of pixels as units instead of length of perturbation vector to measure the perturbation strength and consider the worst case which is one-pixel modification, as well as two other scenarios (i.e. 3 and 5 pixels) for comparison. 

\end{itemize}

%According to the experimental results, the main contributions of our work include:

%!!Include this information in the Section where it is needed!!
%In the following, we abbreviate the originally true class of an adversarial image as ``original class'', the objective class that adversaries desire the DNN to recognize their adversarial images as ``target class'', the DNN classifier that adversaries want to fool as ``target system''. Mainly in Section.4 ,for a specific class $C$, we sometimes call perturbing a data-point of $C$ to another target class as `` going out'' from $C$ and perturbing a data-point of another class to $C$ as ``getting into'' $C$.  

\section{Related works}

The security problem of DNN has become a critical topic  \cite{barreno2010security}\cite {6}.
C. Szegedy et al. first revealed the sensitivity to well-tuned artificial perturbation \cite {3} which can be crafted by several gradient-based algorithms using back-propagation for obtaining gradient information \cite {2}\cite {3}. 
Specifically, I.J.Goodfellow et al. proposed ``fast gradient sign'' algorithm for calculating effective perturbation based on a hypothesis in which the linearity and high-dimensions of inputs are the main reason that a broad class of networks are sensitive to small perturbation \cite {2}. 
S.M. Moosavi-Dezfooli et al. proposed a greedy perturbation searching method by assuming the linearity of DNN decision boundaries \cite {4}. 
In addition, N. Papernot et al. utilize Jacobian matrix to build ``Adversarial Saliency Map'' which indicates the effectiveness of conducting a fixed length perturbation through the direction of each axis  \cite {1}\cite {20}. Except adversarial perturbation, there are other ways of creating adversarial images to make the DNN misclassify, such as artificial image \cite{7} and rotation \cite{103}.
Besides, adversarial perturbation can be also possible in other domains such as speech recognition \cite{100}, natural language processing \cite{101} and malware classification \cite{102}.

A number of detection and defense methods have been also proposed to mitigate the vulnerability induced by adversarial perturbation \cite {106}. For instance, network distillation which was originally proposed for squeezing information of an network to a smaller one is found to be able to reduce the network sensitivity enhancing the robustness of the neural network \cite{107}. Adversarial training \cite{108} is proposed for adding adversarial images to the training data such that the robustness against known adversarial images can be improved. On the other side, some image processing methods are proved to be effective for detecting adversarial images. 
For example, B.Liang et al. show that noise reduction methods such as scalar quantization and spatial smoothing filter can be selectively utilized for mitigating the influence of adversarial perrturbation. By comparing the label of an image before and after the transformation the perturbation can be detected \cite{110}. The method works well on detecting adversarial images with both low and high entropy. Similarly,  W. Xu et al. show that squeezing color bits and local/non-local spatial smoothing can have high success rate on distinguishing adversarial images \cite{109}.
However, recent studies show that many of these defense and detection methods can be effectively evaded by conducting little modification on the original attacks \cite{289, 295, 601}.

Several black-box attacks that require no internal knowledge about the target systems such as gradients, have also been proposed \cite{25}\cite{24}\cite{13}. 
In particular, to the best of our knowledge, the only work before ours that ever mentioned using one-pixel modification to change class labels is carried out by N. Narodytska et al\cite{13}. 
However, differently from our work, they only utilized it as a starting point to derive a further semi black-box attack which needs to modify more pixels (e.g., about 30 pixels out of 1024) without considering the scenario of one-pixel attack. 
%In particular, their one-pixel modification is quite coarse which is almost equivalent to a random search. 
In addition, they have neither measured systematically the effectiveness of the attack nor obtained quantitative results for evaluation. An analysis of the one-pixel attack's geometrical features as well as further discussion about its implications are also lacking. 

There have been many efforts to understand DNN by visualizing the activation of network nodes \cite {19} \cite {18}\cite {16}while the geometrical characteristics of DNN boundary have gained less attraction due to the difficulty of understanding high-dimensional space. 
However, the robustness evaluation of DNN with respect to adversarial perturbation might shed light in this complex problem \cite{27}. 
For example, both natural and random images are found to be vulnerable to adversarial perturbation. 
Assuming these images are evenly distributed, it suggests that most data points in the input space are gathered near to the boundaries \cite{27}. 
In addition, A. Fawzi et al. revealed more clues by conducting a curvature analysis. 
%Their conclusion is that including the connectivity of regions of the same class, 
Their conclusion is that the region along most directions around natural images are flat with only few directions where the space is curved and the images are sensitive to perturbation\cite{26}.  
Interestingly, universal perturbations (i.e. a perturbation that when added to any natural image can generate adversarial images with high effectiveness) were shown possible and to achieve a high effectiveness when compared to random perturbation. 
This indicates that the diversity of boundaries might be low while the boundaries' shapes near different data points are similar \cite{28}. 

\section{Methodology}
\subsection{Problem Description}

Generating adversarial images can be formalized as an optimization problem with constraints. 
We assume an input image can be represented by a vector in which each scalar element represents one pixel.
Let $f$ be the target image classifier which receives n-dimensional inputs, $\textbf{x} = (x_1,..,x_n)$ be the original natural image correctly classified as class $t$. 
The probability of $\textbf{x}$ belonging to the class $t$ is therefore $f_t(\textbf{x})$. 
The vector $e(\textbf{x}) = (e_1,..,e_n)$ is an additive adversarial perturbation according to $\textbf{x}$, the target class $adv$ and the limitation of maximum modification $L$. 
Note that $L$ is always measured by the length of vector $e(\textbf{x})$. 
The goal of adversaries in the case of targeted attacks is to find the optimized solution $e(\textbf{x})^{*}$ for the following question:

\begin{equation*}
\begin{aligned}
& \underset{e(\textbf{x})^{*}}{\text{maximize}}
& & f_{adv}(\textbf{x}+e(\textbf{x})) \\
& \text{subject to}
& & \Vert e(\textbf{x}) \Vert \leq L
%& & \Vert e(\textbf{x}) \Vert_{0} \leq d
\end{aligned}
\end{equation*}

The problem involves finding two values: (a) which dimensions that need to be perturbed and (b) the corresponding strength of the modification for each dimension. 
In our approach, the equation is slightly different:

\begin{equation*}
\begin{aligned}
& \underset{e(\textbf{x})^{*}}{\text{maximize}}
& & f_{adv}(\textbf{x}+e(\textbf{x})) \\
& \text{subject to}
%& & \Vert e(\textbf{x}) \Vert \leq L
& & \Vert e(\textbf{x}) \Vert_{0} \leq d,
\end{aligned}
\end{equation*}
where $d$ is a small number. In the case of one-pixel attack $d=1$.
Previous works commonly modify a part of all dimensions while in our approach only $d$ dimensions are modified with the other dimensions of $e(\textbf{x})$ left to zeros. 
%In our case, the number of dimensions that need to be perturbed are set to constant numbers $1$, $3$ and $5$, in other words, the limitation $L$ is set to be the maximum modification on $1$, $3$ and $5$ pixels. 

%Geometrically, the entire input space of a DNN image classifier can be a high-dimensional cube. 
The one-pixel modification can be seen as perturbing the data point along a direction parallel to the axis of one of the $n$ dimensions. 
Similarly, the 3 (5)-pixel modification moves the data points within 3 (5)-dimensional cubes. 
Overall, few-pixel attack conducts perturbations on the low-dimensional slices of input space. 
%Intuitively, it seems that one-pixel perturbation is merely a small change in the vast high dimensional input space such that it might hardly cause any changes on class label. 
In fact, one-pixel perturbation allows the modification of an image towards a chosen direction out of $n$ possible directions with arbitrary strength. 
	%Therefore, the complete set of all possible adversarial images that can be created from a natural image forms a n-dimensional coordinate inside the input space in which the origin is the image itself. 
%The search for adversarial images starts from the original image and go across the input space through the directions with each of them is perpendicular to any others. 
This is illustrated in Figure~4 for the case when $n=3$. 
%Therefore, even if only modifying one pixel, it allows to search the candidate adversarial images in a fairly wide scope. Compared with previous works using $L_p$ norm to control the overall perturbation strength, which the resulting search spaces are a small sphere around the vicinity of the image, the search of few-pixel attack can go much further in the input space therefore more hopeful to reach other targeted classes.

\begin{figure}[t]
\begin{center}
\includegraphics[width=0.7\linewidth]{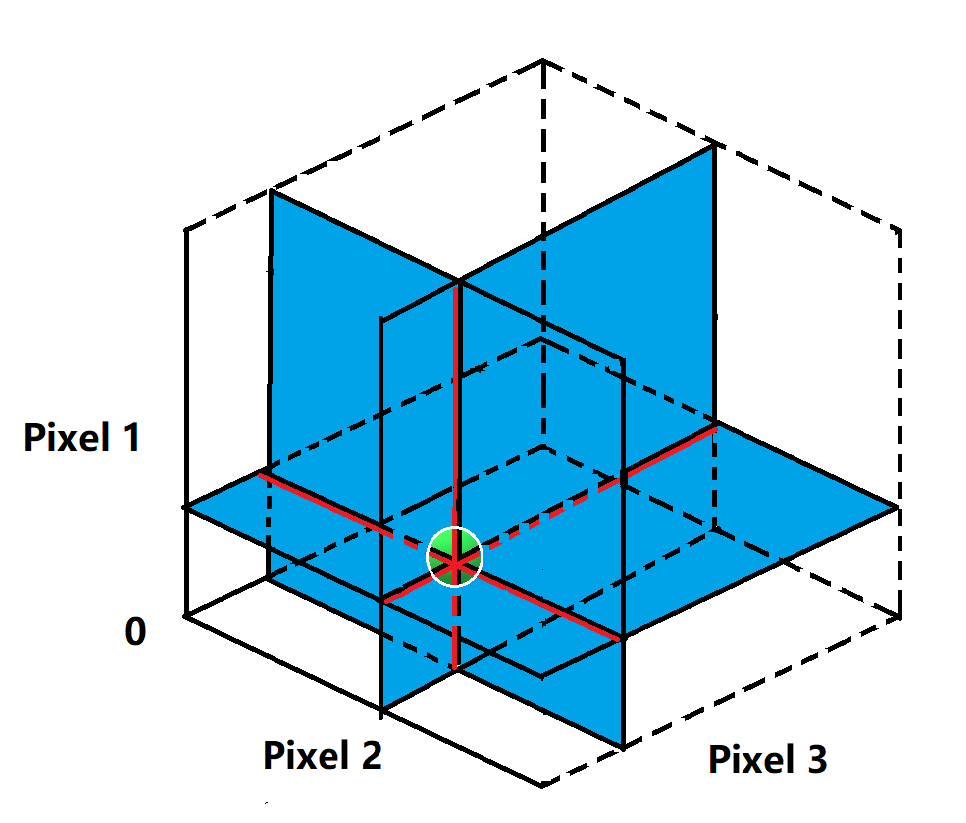}
\end{center}
   \caption{An illustration of using one and two-pixel perturbation attack in a 3-dimensional input space (i.e. the image has three pixels). 
   The green point (sphere) denotes a natural image. In the case of one-pixel perturbation, the search space is the three perpendicular lines that intersect at point of natural image, which are denoted by red and black stripes. 
   For two-pixel perturbation, the search space is the three blue (shaded) two-dimensional planes. 
   In summary, one and two-pixel attacks search the perturbation on respectively one and two dimensional slices of the original three dimensional input space. 
   %In addition, the yellow 3-d sphere indicates the search areas of previous works which utilized $L_p$ norm to control the overall modification strength. Comparatively, our few-pixel attack can search much further areas.
   }
\label{fig:long}
\label{fig:onecol} 
\end{figure}

Thus, usual adversarial images are constructed by perturbating all pixels with an overall constraint on the strength of accumulated modification\cite{29}\cite{28} while the few-pixel attack considered in this paper is the opposite which specifically focus on few pixels but does not limit the strength of modification. 
%Geometrically, the universal perturbation moves small steps in the original input space, while few-pixel attacks use very low dimensional slices to cut the input space and evaluate how class label changes on the resulting cross sections. 

\subsection{Differential Evolution}

Differential evolution (DE) is a population based optimization algorithm for solving complex multi-modal optimization problems \cite{12}, \cite{das2011differential}.
DE belongs to the general class of evolutionary algorithms (EA). 
Moreover, it has mechanisms in the population selection phase that keep the diversity such that in practice it is expected to efficiently find higher quality solutions than gradient-based solutions or even other kinds of EAs \cite{civicioglu2013conceptual}.
In specific, during each iteration another set of candidate solutions (children) is generated according to the current population (parents). Then the children are compared with their corresponding parents, surviving if they are more fitted (possess higher fitness value) than their parents. 
In such a way, only comparing the parent and his child, the goal of keeping diversity and improving fitness values can be simultaneously achieved. 

%DE does not use the gradient information for optimizing therefore it does not require the objective function to be differentiable.
DE does not use the gradient information for optimizing and therefore does not require the objective function to be differentiable or previously known.
Thus, it can be utilized on a wider range of optimization problems compared to gradient based methods (e.g., non-differentiable, dynamic, noisy, among others). 
The use of DE for generating adversarial images have the following main advantages:

\begin{itemize}

\item \textbf{Higher probability of Finding Global Optima} - DE is a meta-heuristic which is relatively less subject to local minima than gradient descent or greedy search algorithms (this is in part due to diversity keeping mechanisms and the use of a set of candidate solutions). Moreover, the problem considered in this article has a strict constraint (only one pixel can be modified) making it relatively harder.

\item \textbf{Require Less Information from Target System} -  DE does not require the optimization problem to be differentiable as is required by classical optimization methods such as gradient descent and quasi-newton methods. 
This is critical in the case of generating adversarial images since 1) There are networks that are not differentiable, for instance \cite {5}.  
2) Calculating gradient requires much more information about the target system which can be hardly realistic in many cases. 
%Although training an approximated model through the black-box feedback from the target is theoretically possible, the accuracy of approximation can be influenced by many factors in practice while time and resource consuming. 
%Moreover, using DE requires neither gradient information nor training any additional models. 
%It can directly craft adversarial images based on only the black-box feedback. 

\item \textbf{Simplicity} - The approach proposed here is independent of the classifier used. 
For the attack to take place it is sufficient to know the probability labels.
%Most of previous works abstract the problem of searching the effective perturbation to specific optimization problem (e.g. an explicit target function with constraints). Namely they made additional assumptions to the searching problem and might bring additional complexity. Our method does not solve any explicit target functions but directly works with the probability label value of the target classes.
\end{itemize}

There are many DE variations/improvements such as self-adaptive \cite{brest2006self},  multi-objective \cite{vargas2015general}, among others.
The current work can be further improved by taking these variations/improvements into account.

\subsection{Method and Settings}
We encode the perturbation into an array (candidate solution) which is optimized (evolved) by differential evolution. 
One candidate solution contains a fixed number of perturbations and each perturbation is a tuple holding five elements: x-y coordinates and RGB value of the perturbation. 
One perturbation modifies one pixel. 
The initial number of candidate solutions (population) is $400$ and at each iteration another $400$ candidate solutions (children) will be produced by using the usual DE formula:

\begin{eqnarray*}
%\begin{aligned}
& x_i(g + 1) = x_{r1}(g) + F(x_{r2}(g) - x_{r3}(g)),\\
& r1\not = r2\not = r3,
%\end{aligned}
\end{eqnarray*}

where $x_i$ is an element of the candidate solution, $r1, r2, r3$ are random numbers, $F$ is the scale parameter set to be 0.5, $g$ is the current index of generation.
Once generated, each candidate solution compete with their corresponding parents according to the index of the population and the winner survive for next iteration. 
The maximum number of iteration is set to $100$ and early-stop criterion will be triggered when the probability label of target class exceeds $90\%$ in the case of targeted attacks on Kaggle CIFAR-10, and when the label of true class is lower than $5\%$ in the case of non-targeted attacks on ImageNet. Then the label of true class is compared with the highest non-true class to evaluate if the attack succeeded. The initial population is initialized by using uniform distributions $U (1, 32)$ for CIFAR-10 images and $U (1, 227)$ for ImageNet images, for generating x-y coordinate (e.g., the image has a size of 32X32 in CIFAR-10 and for ImageNet we unify the original images with various resolutions to 227X227) and Gaussian distributions N ($\mu$=128, $\sigma$=127) for RGB values. The fitness function is simply the probabilistic label of the target class in the case of CIFAR-10 and the label of true class in the case of ImageNet. The crossover is not included in our scheme.

\section{Evaluation and Results}
The evaluation of the proposed attack method is based on CIFAR-10 and ImageNet datasets. We introduce several metrics to measure the effectiveness of the attacks:

\begin{itemize}

\item \textbf{Success Rate} - In the case of non-targeted attacks, it is defined as the percentage of adversarial images that were successfully classified by the target system as an arbitrary target class. In the case of targeted attack, it is defined as the probability of perturbing a natural image to a specific target class.

\item \textbf{Adversarial Probability Labels (Confidence)} - Accumulates the values of probability label of the target class for each successful perturbation, then divided by the total number of successful perturbations. The measure indicates the average confidence given by the target system when mis-classifying adversarial images.

\item \textbf{Number of Target Classes} - Counts the number of natural images that successfully perturb to a certain number (i.e. from 0 to 9) of target classes. In particular, by counting the number of images that can not be perturbed to any other classes, the effectiveness of non-targeted attack can be evaluated.

\item \textbf{Number of Original-Target Class Pairs} - Counts the number of times each original-destination class pair was attacked.

\end{itemize}

\subsection{Kaggle CIFAR-10}
We train 3 types of common networks: All convolution network \cite {21}, Network in Network\cite {31} and VGG16 network\cite {32} as target image classifiers on CIFAR-10 dataset \cite {11,703}. The structures of the networks are described in Table~ 1, 2 and 3. 
The network setting were kept as similar as possible to the original with a few modifications in order to get the highest classification accuracy. 
Both the scenarios of targeted and non-targeted attacks are considered.
For each of the attacks on the three types of neural networks $500$ natural images are randomly selected from the Kaggle CIFAR-10 test dataset to conduct the attack. 

Note that we use the Kaggle CIFAR-10 test dataset \cite{703} instead of the original one for this experiments. The dataset contains 300,000 cifar-10 images which can be visually inspected to have the following modifications: duplication, rotation, clipping, blurring, adding few random bad pixels and so on. However, the exact employed modification algorithm is not released. This makes it a more practical dataset which simulates common scenarios that images can contain unknown random noise. We also show the results on the original CIFAR-10 test dataset in Section 5 for comparison.

In addition, an experiment is conducted on the all convolution network  \cite {21} by generating $500$ adversarial images with three and five pixel-modification. 
The objective is to compare one-pixel attack with three and five pixel attacks. 
For each natural image, nine target attacks are launched trying to perturb it to the other 9 target classes. Note that we actually only launch targeted attacks and the effectiveness of non-targeted attack is evaluated based on targeted attack results. That is, if an image can be perturbed to at least one target class out of total 9 classes, the non-targeted attack on this image succeeds.
Overall, it leads to the total of $36000$ adversarial images created. 
To evaluate the effectiveness of the attacks, some established measures from the literature are used as well as some new kinds of measures are introduced: 

\begin{table}
\begin{center}
\begin{tabular}{|c|c|}
\hline
conv2d layer(kernel=3, stride = 1, depth=96) \\
conv2d layer(kernel=3, stride = 1, depth=96) \\
conv2d layer(kernel=3, stride = 2, depth=96) \\
conv2d layer(kernel=3, stride = 1, depth=192) \\
conv2d layer(kernel=3, stride = 1, depth=192) \\
dropout(0.3) \\
conv2d layer(kernel=3, stride = 2, depth=192) \\
conv2d layer(kernel=3, stride = 2, depth=192) \\
conv2d layer(kernel=1, stride = 1, depth=192) \\
conv2d layer(kernel=1, stride = 1, depth=10) \\
average pooling layer(kernel=6, stride=1) \\
flatten layer \\
softmax classifier \\
\hline
\end{tabular}
\end{center}
\caption{All convolution network}

\begin{center}
\begin{tabular}{|c|c|}
\hline
conv2d layer(kernel=5, stride = 1, depth=192) \\
conv2d layer(kernel=1, stride = 1, depth=160) \\
conv2d layer(kernel=1, stride = 1, depth=96) \\
max pooling layer(kernel=3, stride=2) \\
dropout(0.5) \\
conv2d layer(kernel=5, stride = 1, depth=192) \\
conv2d layer(kernel=5, stride = 1, depth=192) \\
conv2d layer(kernel=5, stride = 1, depth=192) \\
average pooling layer(kernel=3, stride=2) \\
dropout(0.5) \\
conv2d layer(kernel=3, stride = 1, depth=192) \\
conv2d layer(kernel=1, stride = 1, depth=192) \\
conv2d layer(kernel=1, stride = 1, depth=10) \\
average pooling layer(kernel=8, stride=1) \\
flatten layer \\
softmax classifier \\
\hline
\end{tabular}
\end{center}
\caption{Network in Network}

\begin{center}
\begin{tabular}{|c|c|}
\hline
conv2d layer(kernel=3, stride = 1, depth=64) \\
conv2d layer(kernel=3, stride = 1, depth=64) \\
max pooling layer(kernel=2, stride=2) \\
conv2d layer(kernel=3, stride = 1, depth=128) \\
conv2d layer(kernel=3, stride = 1, depth=128) \\
max pooling layer(kernel=2, stride=2) \\
conv2d layer(kernel=3, stride = 1, depth=256) \\
conv2d layer(kernel=3, stride = 1, depth=256) \\
conv2d layer(kernel=3, stride = 1, depth=256) \\
max pooling layer(kernel=2, stride=2) \\
conv2d layer(kernel=3, stride = 1, depth=512) \\
conv2d layer(kernel=3, stride = 1, depth=512) \\
conv2d layer(kernel=3, stride = 1, depth=512) \\
max pooling layer(kernel=2, stride=2) \\
conv2d layer(kernel=3, stride = 1, depth=512) \\
conv2d layer(kernel=3, stride = 1, depth=512) \\
conv2d layer(kernel=3, stride = 1, depth=512) \\
max pooling layer(kernel=2, stride=2) \\
flatten layer \\
fully connected(size=2048) \\
fully connected(size=2048) \\
softmax classifier \\
\hline
\end{tabular}
\end{center}
\caption{VGG16 network}

\end{table}

\subsection{ImageNet}
For ImageNet we applied a non-targeted attack with the same DE parameter settings used on the CIFAR-10 dataset, although ImageNet has a search space 50 times larger than CIFAR-10. Note that we actually launch the non-targeted attack for ImageNet by using a fitness function that aims to decrease the probability label of the true class. Different from CIFAR-10, whose effectiveness of non-targeted attack is calculated based on the targeted attack results carried out by using a fitness function for increasing the probability of target classes. 
Given the time constraints, we conduct the experiment without proportionally increasing the number of evaluations, i.e. we keep the same number of evaluations. Our tests are run over the BVLC AlexNet using 105 images from ILSVRC 2012 test set selected randomly for the attack. For ImageNet we only conduct one pixel attack because we want to verify if such a tiny modification can fool images with larger size and if it is computationally tractable to conduct such attacks.
The ILSVRC 2012 images are in lossy jpeg format with non-unified sizes. In order to reduce the practical interference to the evaluation as much as possible, we first convert all target images from jpeg to png therefore during later processing it will be lossless. The images are further resized to 227X227 resolution for inputting to AlexNet (using nearest filter). Then we follow the same procedure to attacking CIFAR-10. Note that the discrepancy on pre-processing raw images (e.g., using center cropping instead of simple resizing) can influence the classification performance of AlexNet and attack rate. 
Here we only show the result on one setting and leave the comprehensive evaluation of attacking AlexNet using difference pre-processing methods for future work.

\subsection{Results}
The success rates and adversarial probability labels for one-pixel perturbations on three CIFAR-10 networks and BVLC network are shown in Table~4 and the three and five-pixel perturbations on Kaggle CIFAR-10 is shown in Table~5.  
The number of target classes is shown by Figure~5. 
The number of original-target class pairs is shown by the heat-maps of Figure~6 and 7. 
In addition to the number of original-target class pairs, the total number of times each class had an attack which either originated or targeted it is shown in Figure~8.
Since only non-targeted attacks are launched on ImageNet, the ``Number of target classes'' and ``Number of original-target class pairs'' metrics are not included in the ImageNet results.

\subsubsection{Success Rate and Adversarial Probability Labels (Targeted Attack Results)}
On Kaggle CIFAR-10, the success rates of one-pixel attacks on three types of networks show the generalized effectiveness of the proposed attack through different network structures. 
On average, each image can be perturbed to about two target classes for each network. 
In addition, by increasing the number of pixels that can be modified to three and five, the number of target classes that can be reached increases significantly. 
By dividing the adversarial probability labels by the success rates, the confidence values (i.e. probability labels of target classes) are obtained which are 79.39$\%$, 79.17$\%$ and 77.09$\%$ respectively to one, three and five-pixel attacks. 

On ImageNet, the results show that the one pixel attack generalizes well to large size images and fool the corresponding neural networks. In particular, there is $16.04\%$ chance that an arbitrary ImageNet test image can be perturbed to a target class with $22.91\%$ confidence. Note that the ImageNet results are done with the same settings as CIFAR-10 while the resolution of images we use for the ImageNet test is 227x227, which is 50 times larger than CIFAR-10 (32x32). 
Notice that in each successful attack the probability label of the target class is the highest. Therefore, the confidence of $22.91\%$ is relatively low but tell us that the other remaining $999$ classes are even lower to an almost uniform soft label distribution.  Thus, the one-pixel attack can break the confidence of BVLC AlexNet to a nearly uniform soft label distribution. The low confidence is caused by the fact that we utilized a non-targeted evaluation that only focuses on decreasing the probability of the true class. Other fitness functions should give different results.

\begin{table}
\begin{center}
\begin{tabular}{|c|c|c|c|c|}
	\hline  &AllConv&NiN&VGG16&BVLC\\
	\hline OriginAcc&$85.6\%$&$87.2\%$&$83.3\%$&$57.3\%$\\
	\hline Targeted&$19.82\%$&$23.15\%$&$16.48\%$&--\\
	\hline Non-targeted&$68.71\%$&$71.66\%$&$63.53\%$&$16.04\%$\\
	%\hline Accumulated labels&$23.08\%$&$25.29\%$&$19.37\%$&$--\%$\\
	\hline Confidence&$79.40\%$&$75.02\%$&$67.67\%$&$22.91\%$\\
	\hline
\end{tabular}
\end{center}
\caption{Results of conducting one-pixel attack on four different types of networks: All Convolutional network (AllConv), Network in Network (NiN), VGG16 and BVLC AlexNet. The OriginalAcc is the accuracy on the natural test datasets. Targeted/Non-targeted indicate the accuracy of conducting targeted/non-targeted attacks. Confidence is the average probability of target classes.}
\end{table}

\begin{table}
\begin{center}
\begin{tabular}{|c|c|c|}
	\hline  &3 pixels &5 pixels\\
	\hline Success rate(tar)&$40.57\%$&$44.00\%$\\
	\hline Success rate(non-tar)&$86.53\%$&$86.34\%$\\
	%\hline Accumulated labels&$35.08\%$&$42.36\%$\\
	\hline Rate/Labels&$79.17\%$&$77.09\%$\\
	\hline
\end{tabular}
\end{center}
\caption{Results of conducting three-pixel attack on AllConv networks and five-pixel attack on Network in network.}
\end{table}

\subsubsection{Number of Target Classes (Non-targeted Attack Results)}

\begin{figure}[t]
\begin{center}
\includegraphics[width=0.8\linewidth]{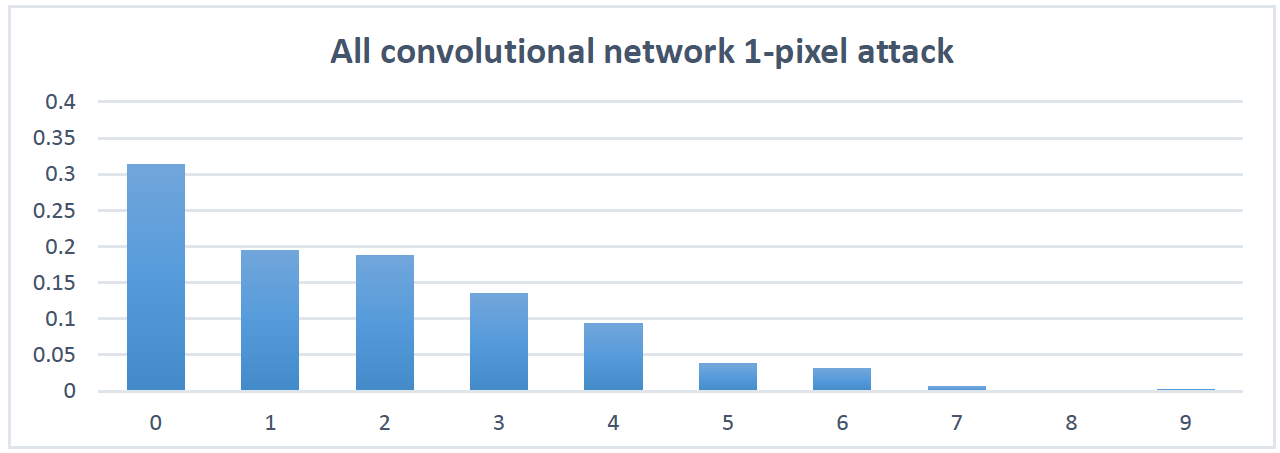}
\end{center}
\begin{center}
\includegraphics[width=0.8\linewidth]{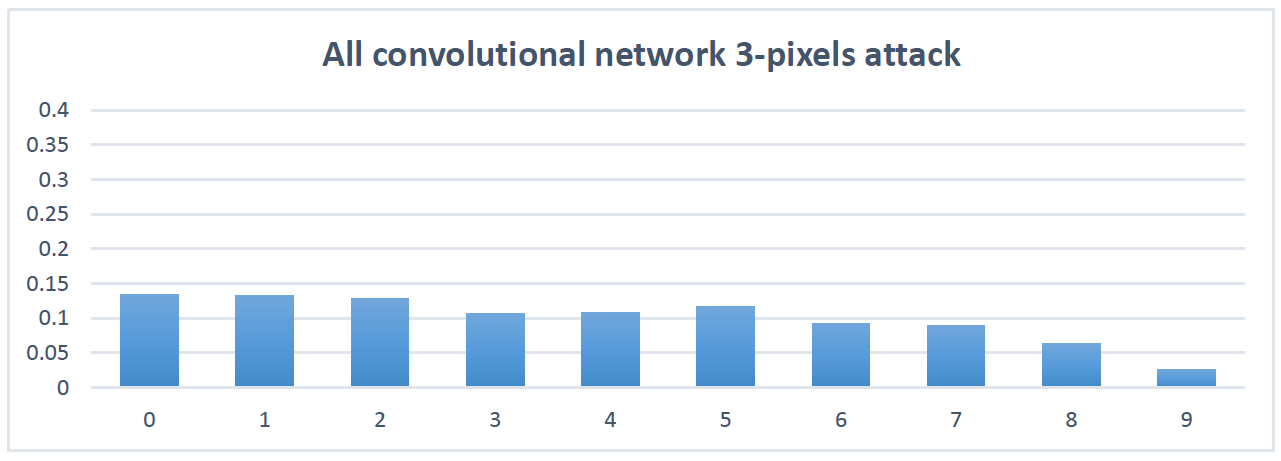}
\end{center}
\begin{center}
\includegraphics[width=0.8\linewidth]{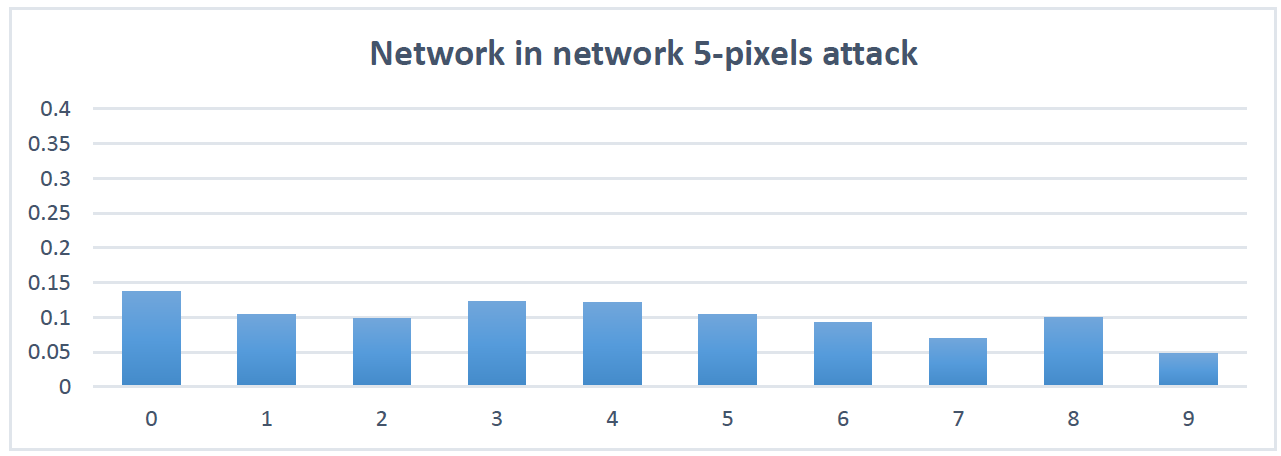}
\end{center}
\begin{center}
\includegraphics[width=0.8\linewidth]{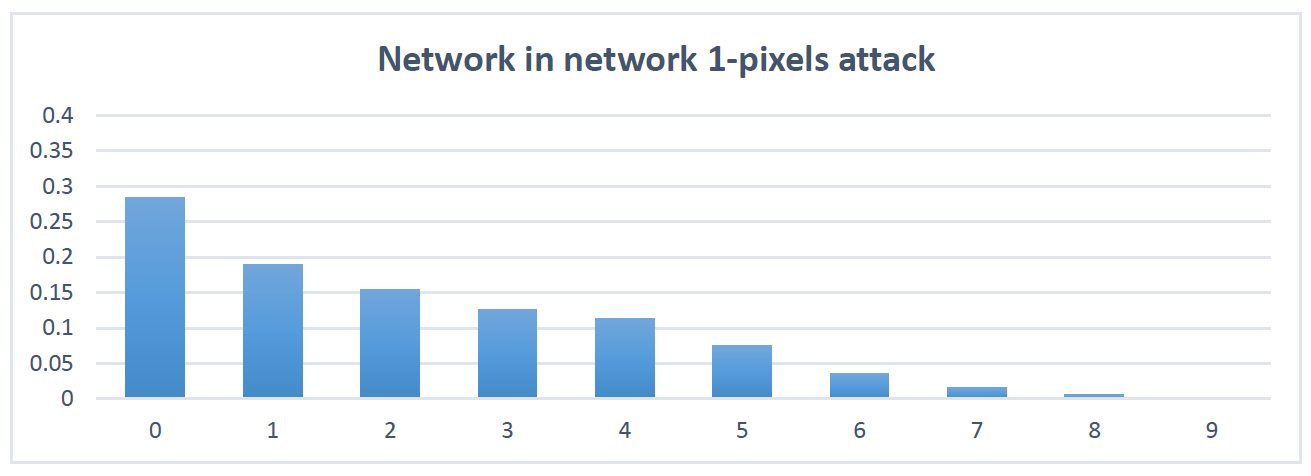}
\end{center}
\begin{center}
\includegraphics[width=0.8\linewidth]{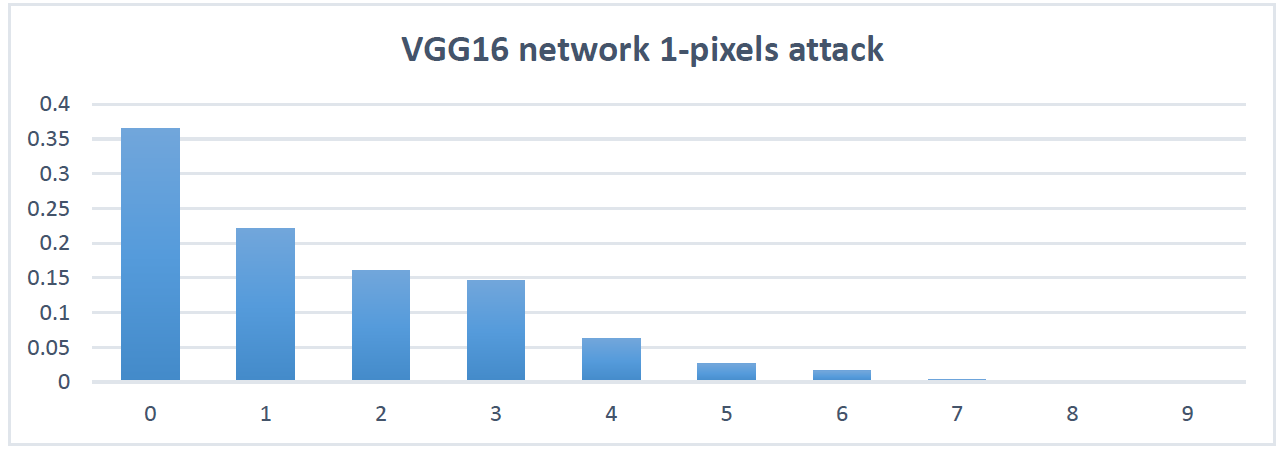}
\end{center}
   \caption{The graphs shows the percentage of natural images that were successfully perturbed to a certain number (from 0 to 9) of target classes by using one, three or five-pixel perturbation. 
   The vertical axis shows the percentage of images that can be perturbed while the horizontal axis indicates the number of target classes.}
\label{fig:long}
\label{fig:onecol} 
\end{figure}

\begin{figure}[t]
\begin{center}
\includegraphics[width=0.6\linewidth]{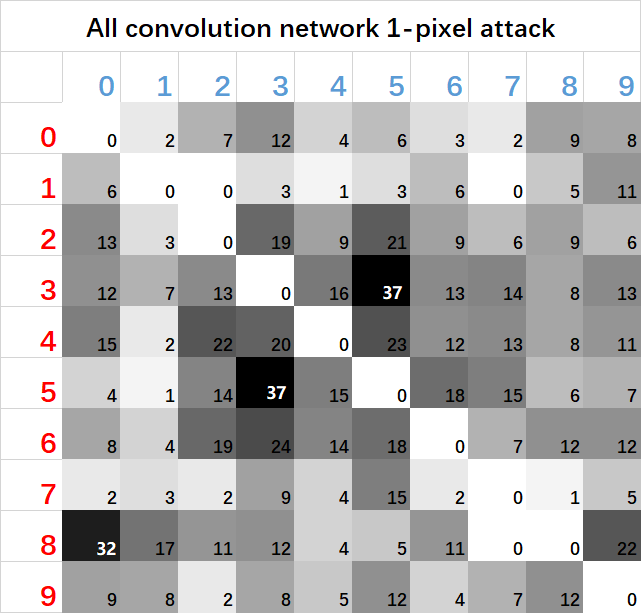}
\end{center}
  
\begin{center}
\includegraphics[width=0.6\linewidth]{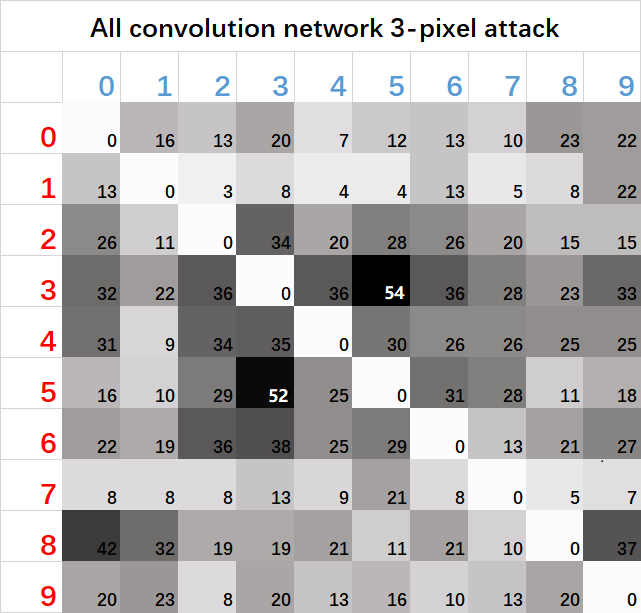}
\end{center}

\begin{center}
\includegraphics[width=0.6\linewidth]{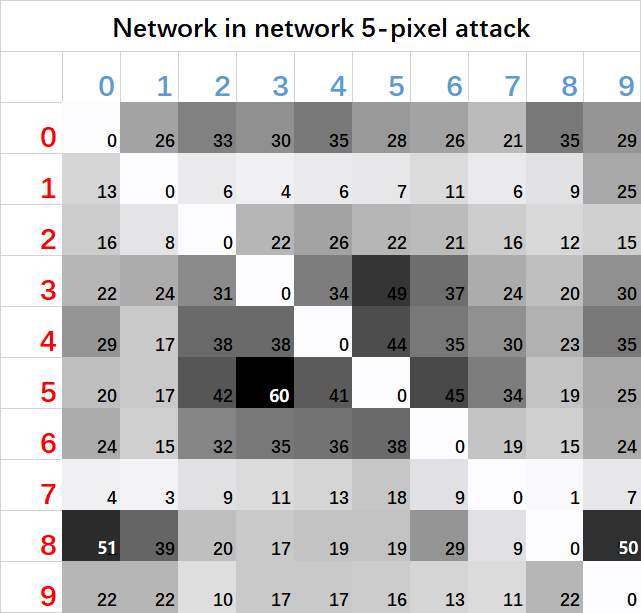}
\end{center}

\caption{Heat-maps of the number of times a successful attack is present with the corresponding original-target class pair in one, three and five-pixel attack cases. 
Red (vertical) and (horizontal) blue indices indicate respectively the original and target classes. 
The number from $0$ to $9$ indicates respectively the following classes: airplane, automobile, bird, cat, deer, dog, frog, horse, ship, truck.}
\label{fig:long}
\label{fig:onecol}
\end{figure}

\begin{figure}[t]
\begin{center}
\includegraphics[width=0.6\linewidth]{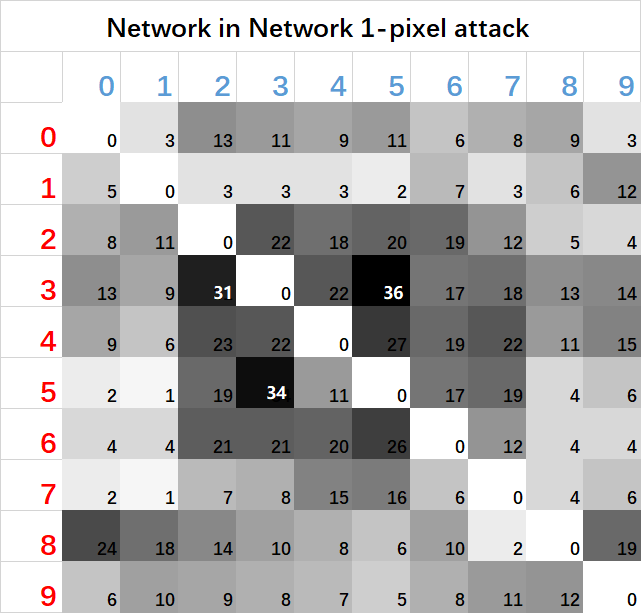}
\end{center}

\begin{center}
\includegraphics[width=0.6\linewidth]{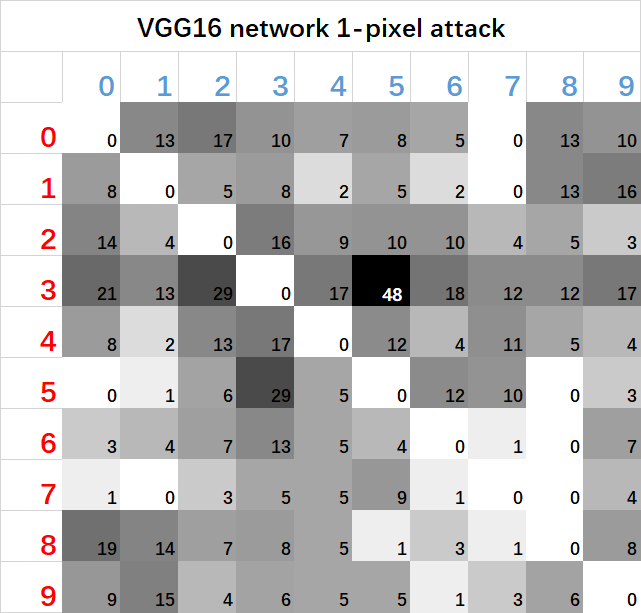}
\end{center}

   \caption{Heat-maps for one-pixel attack on Network in network and VGG.}
\label{fig:long}
\label{fig:onecol}
\end{figure}

\begin{figure}[t]
\begin{center}
\includegraphics[width=1.0\linewidth]{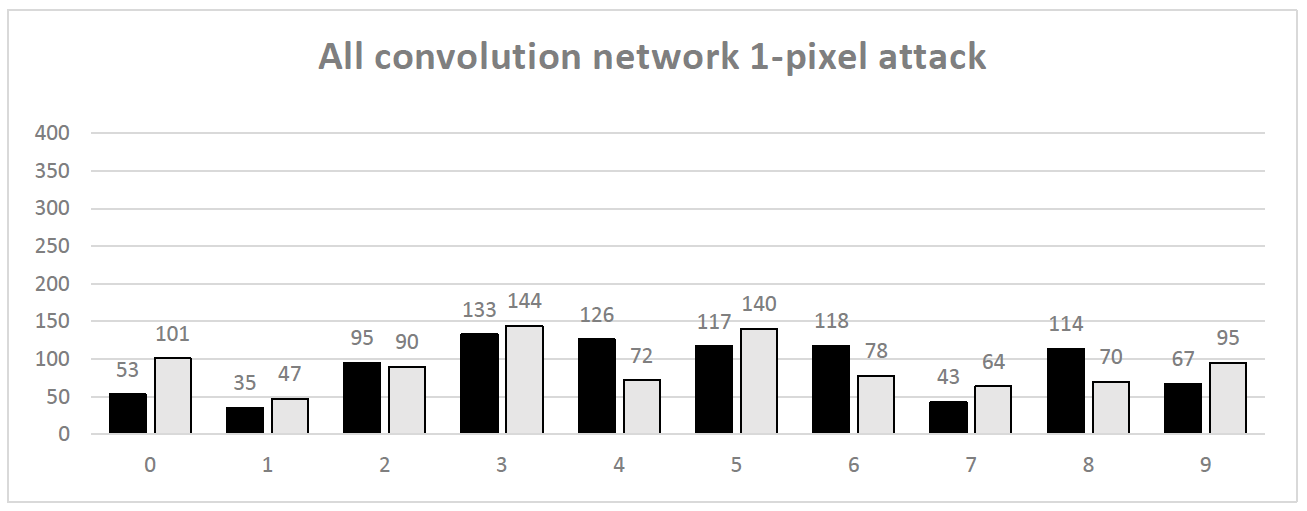}
\end{center}
\begin{center}
\includegraphics[width=1.0\linewidth]{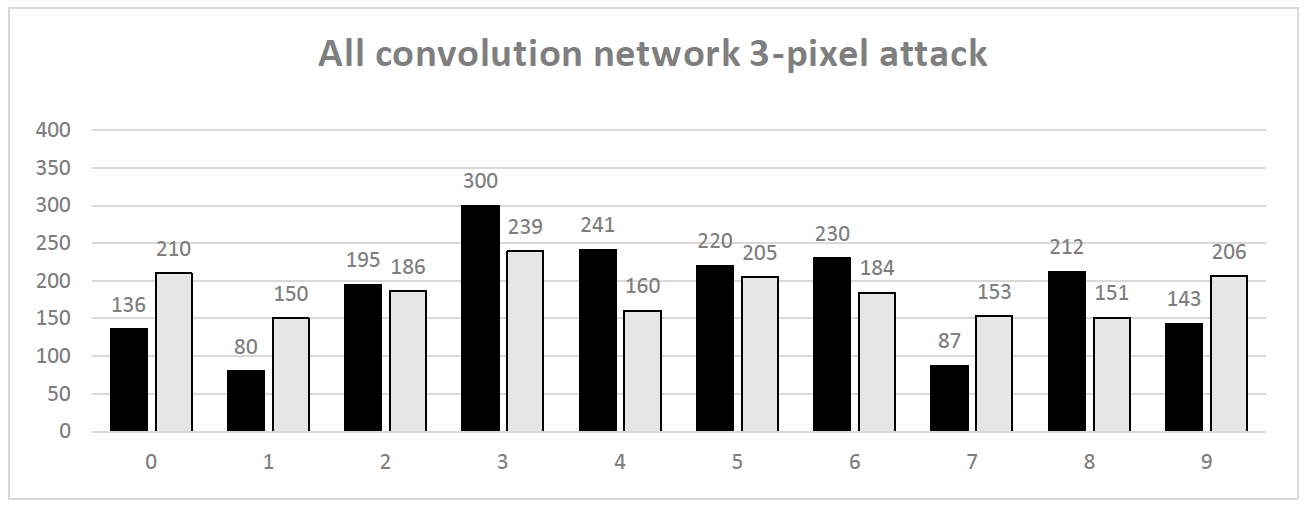}
\end{center}
\begin{center}
\includegraphics[width=1.0\linewidth]{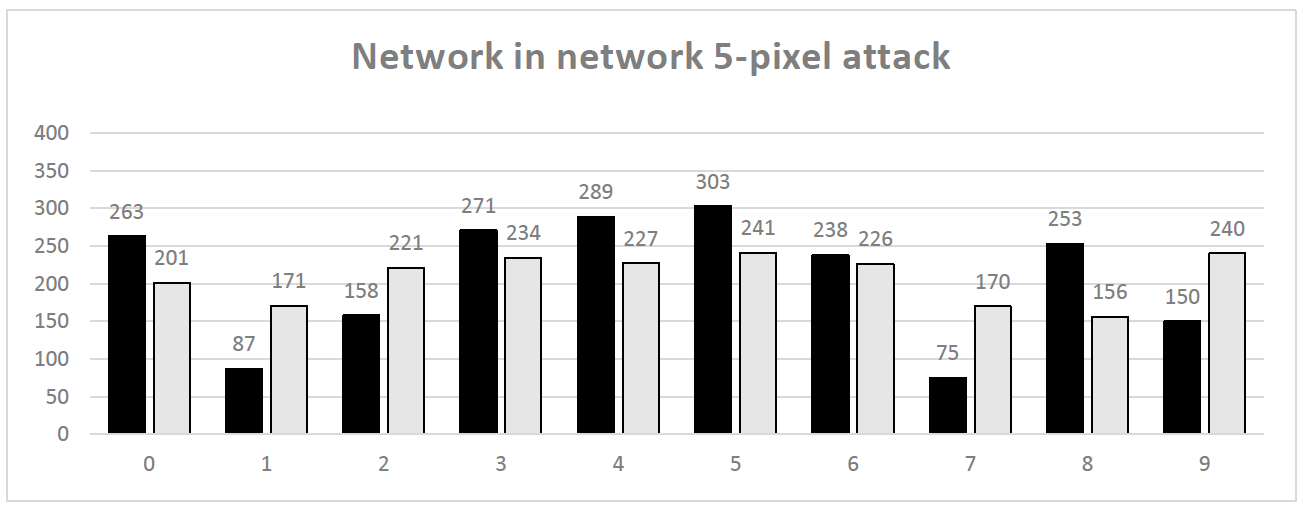}
\end{center}
\begin{center}
\includegraphics[width=1.0\linewidth]{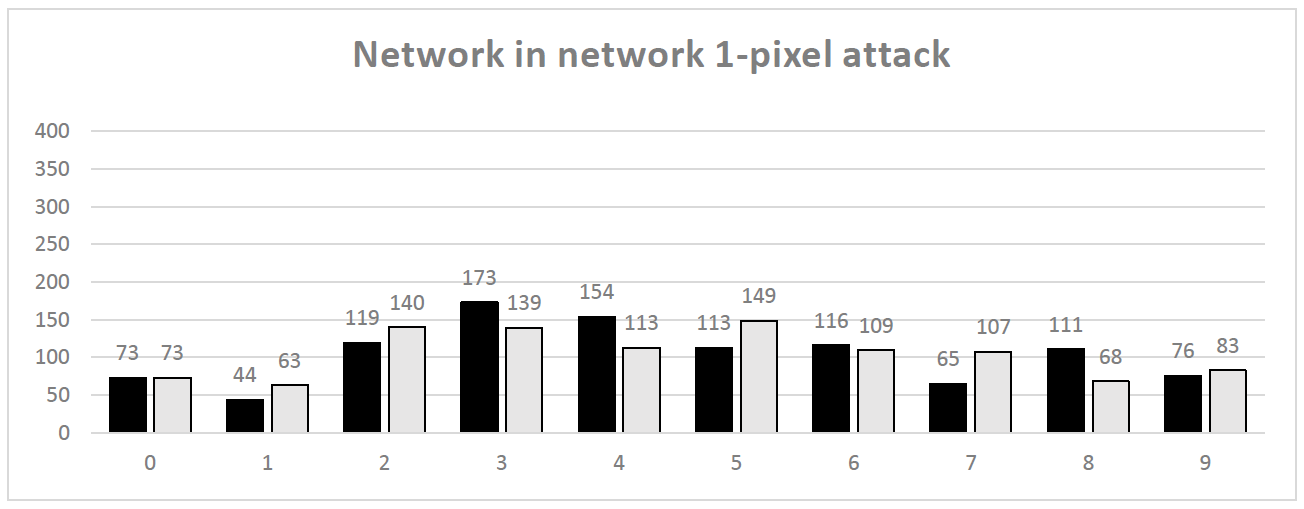}
\end{center}
\begin{center}
\includegraphics[width=1.0\linewidth]{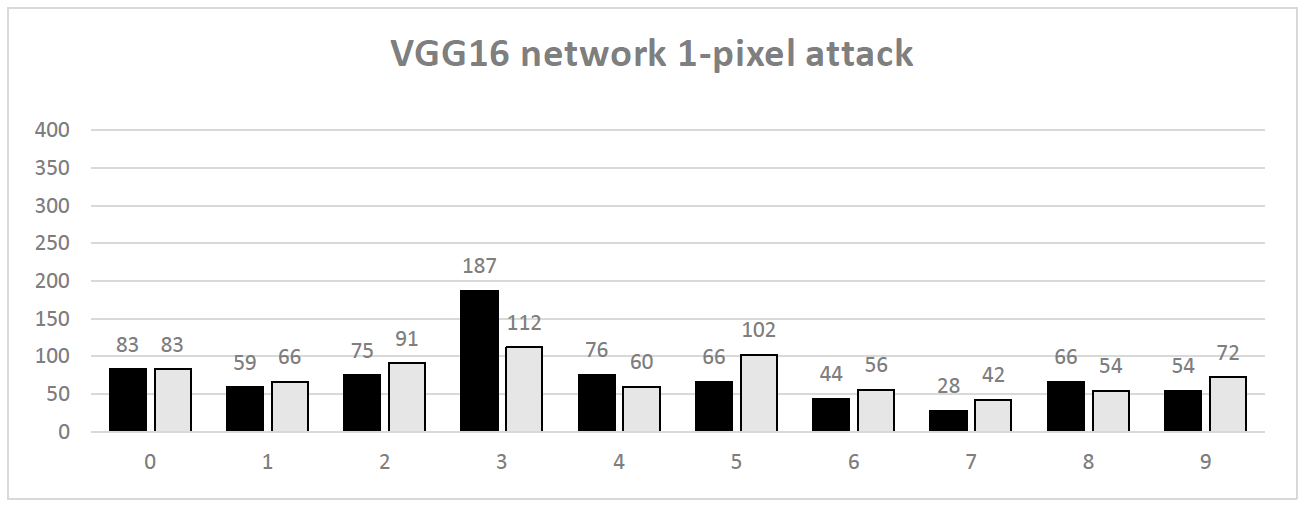}
\end{center}
   \caption{Number of successful attacks (vertical axis) for a specific class acting as the original (black) and target (gray) class. 
   The horizontal axis indicates the index of each class which is the same as Figure~7.}
\label{fig:long}
\label{fig:onecol} 
\end{figure}

\begin{figure}[t]
\begin{center}
\includegraphics[width=0.8\linewidth]{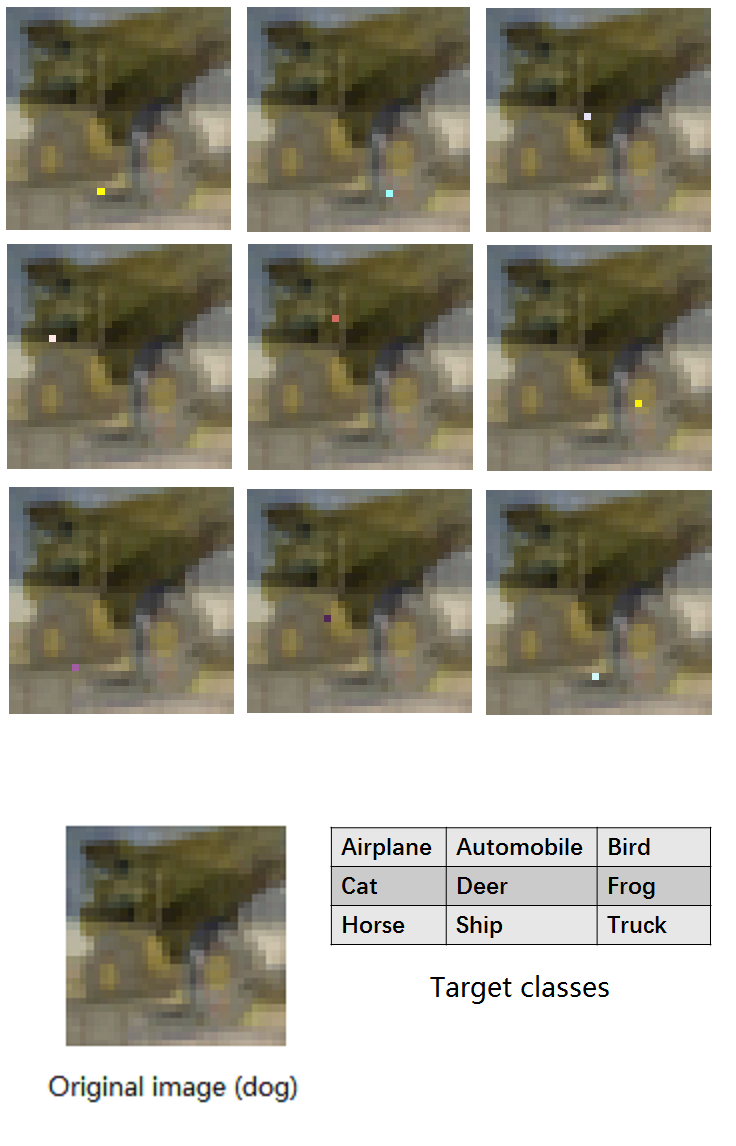}
\end{center}
   \caption{A natural image of the dog class that can be perturbed to all other nine classes. 
   The attack is conducted over the AllConv network using the proposed one pixel attack. 
   The table in the bottom shows the class labels output by the target DNN, all with approximately 100$\%$ confidence. 
   This curious result further emphasize the difference and limitations of current methods when compared to human recognition.}
   %ere such small pertubations would not cause a dog to be confused with a truck.}
\label{fig:long}
\label{fig:onecol} 
\end{figure}

Regarding the results shown in Figure~5, we find that with only one-pixel modification a fair amount of natural images can be perturbed to two, three and four target classes. 
By increasing the number of pixels modified, perturbation to more target classes becomes highly probable. 
In the case of non-targeted one-pixel attack, the VGG16 network got a slightly higher robustness against the proposed attack. 
This suggests that all three types of networks (AllConv network, NiN and VGG16) are vulnerable to this type of attack.

The results of attacks are competitive with previous non-targeted attack methods which need much more distortions (Table~6).
It shows that using one dimensional perturbation vectors is enough to find the corresponding adversarial images for most of the natural images. 
In fact, by increasing the number of pixels up to five, a considerable number of images can be simultaneously perturbed to eight target classes. 
In some rare cases, an image can go to all other target classes with one-pixel modification, which is illustrated in Figure~9.
\subsubsection{Original-Target Class Pairs}
Some specific original-target class pairs are much more vulnerable than others (Figure~6 and~7).
For example, images of cat (class 3) can be much more easily perturbed to dog (class 5) but can hardly reach the automobile (class 1). 
This indicates that the vulnerable target classes (directions) are shared by different data points that belong to the same class. 
Moreover, in the case of one-pixel attack, some classes are more robust than others since their data points can be relatively hard to perturb to other classes. 
Among these data points, there are points that can not be perturbed to any other classes. 
This indicates that the labels of these points rarely change when going across the input space through $n$ directions perpendicular to the axes.
Therefore, the corresponding original classes are kept robust along these directions. 
However, it can be seen that such robustness can rather easily be broken by merely increasing the dimensions of perturbation from one to three and five because both success rates and number of target classes that can be reached increase when conducting higher-dimensional perturbations.

Additionally, it can also be seen that each heat-map matrix is approximately symmetric, indicating that each class has similar number of adversarial images which were crafted from these classes as well as to these classes (Figure~8).
Having said that, there are some exceptions for example the class 8 (ship) when attacking NiN, the class 4 (deer) when attacking AllConv networks with one pixel, among others.
In the ship class when attacking NiN networks, for example, it is relatively easy to craft adversarial images from them while it is relatively hard to craft adversarial images to them. 
Such unbalance is intriguing since it indicates the ship class is similar to most of the other classes like truck and airplane but not vice-versa. 
This might be due to (a) boundary shape and (b) how close are natural images to the boundary.
In other words, if the boundary shape is wide enough it is possible to have natural images far away from the boundary such that it is hard to craft adversarial images from it.
On the contrary, if the boundary shape is mostly long and thin with natural images close to the border, it is easy to craft adversarial images from them but hard to craft adversarial images to them.

In practice, such classes which are easy to craft adversarial images from may be exploited by malicious users which may make the whole system vulnerable.
In the case here, however, the exceptions are not shared between the networks, revealing that whatever is causing the phenomenon is not shared.
Therefore, for the current systems under the given attacks, such a vulnerability seems hard to be exploited.
%because different networks seems to exhibit different behaviors related to these easily exploited classes.

\subsubsection{Time complexity and average distortion}

To evaluate the time complexity we use the number of evaluations which is a common metric in optimization. In the DE case the number of evaluations is equal to the population size multiplied by the number of generations. We also calculate the average distortion on the single pixel attacked by taking the average modification on the three color channels, which is a more straight forward and explicit measure of modification strength. We did not use the $L_p$ norm due to its limited effectiveness of measuring perceptiveness \cite{105}. The results of two metrics are shown in Table~7.

\begin{table}
\begin{center}
\begin{tabular}{|c|c|c|c|c|}
	\hline  &AllConv&NiN&VGG16&BVLC\\     
	\hline AvgEvaluation&16000&12400&20000&25600\\  %times the populations        
	\hline AvgDistortion&123&133&145&158\\
	\hline
\end{tabular}
\end{center}
\caption{Cost of conducting one-pixel attack on four different types of networks. AvgEvaluation is the average number of evaluations to produce adversarial images. AvgDistortion is the required average distortion in one-channel of a single pixel to produce adversarial images.}
\end{table}

\subsubsection{Comparing with Random One-Pixel Attack}

We compare the proposed method with the random attack to evaluate if DE is truly helpful for conducting one-pixel non-targeted attack on Kaggle CIFAR-10 dataset, which is shown in Table~8. 

Specifically, for each natural image, the random search repeats 100 times, each time randomly modifies one random pixel of the image with random RGB value to attempt to change its label. The confidence of the attack with respect to one image is set to be the highest probability target class label of 100 attacks.

\begin{table}
\begin{center}
\begin{tabular}{|c|c|c|c|c|}
	\hline  &AllConv&NiN&VGG16\\
	\hline DE success rate&$68.71\%$&$71.66\%$&$63.53\%$\\
	\hline Confidence&$79.40\%$&$75.02\%$&$67.67\%$\\
	\hline Random Search success rate&$49.70\%$&$41.72\%$&$15.57\%$\\
	\hline Confidence&$87.73\%$&$75.83\%$&$59.90\%$\\
	\hline
\end{tabular}
\end{center}
\caption{A comparison of attack rate and confidence between DE one-pixel attack and random one-pixel attack (Non-targeted) on Kaggle CIFAR-10 dataset.}
\end{table}

In this experiment, we use the same number of evaluations (80000) for both DE
and random search. According to the comparison, the DE is superior to the random attack regarding attack accuracy, especially in the case of VGG16 network. Specifically, DE is $19.01\%$, $29.94\%$ and $47.96\%$ more efficient than random search respectively for All Convolutional Network, Network in Network and VGG16.
Even with a less efficient result, random search is shown to find $49.70\%$ and $41.72\%$ of the time for respectively All Convolutional Network and Network in Network, therefore the vulnerable pixels that can change the image label significantly are quite common. That seems not to be the case for VGG though in which random search achieves only $15.57\%$. DE has a similar accuracy in all of them showing also a better robustness.

\subsubsection{Change in fitness values}

We run an experiment over different networks to examine how the fitness changes during evolution. The 30 (15) curves come from 30 (15) random Kaggle CIFAR-10 (ImageNet) images successfully attacked by the proposed one-pixel attack (Figure~10). The fitness values are, as previously described, set to be the probability label of the true class for each image. The goal of the attack is to minimize this fitness value.
According to the results, it can be seen that the fitness values can occasionally drop abruptly between two generations while in other cases they decrease smoothly. Moreover, the average fitness value decreases monotonically with the number of generations, showing that the evolution works as expected. We also find that BVLC network is harder to fool due to the smaller decrease in fitness values.

\begin{figure}[h]
\begin{center}
\includegraphics[width=1.0\linewidth]{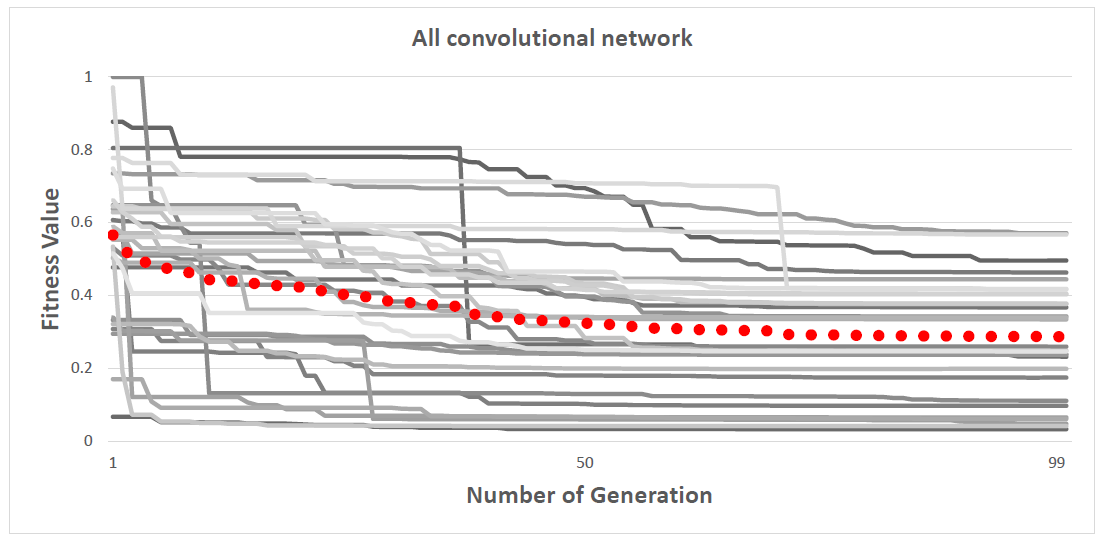}

\end{center}
\begin{center}
\includegraphics[width=1.0\linewidth]{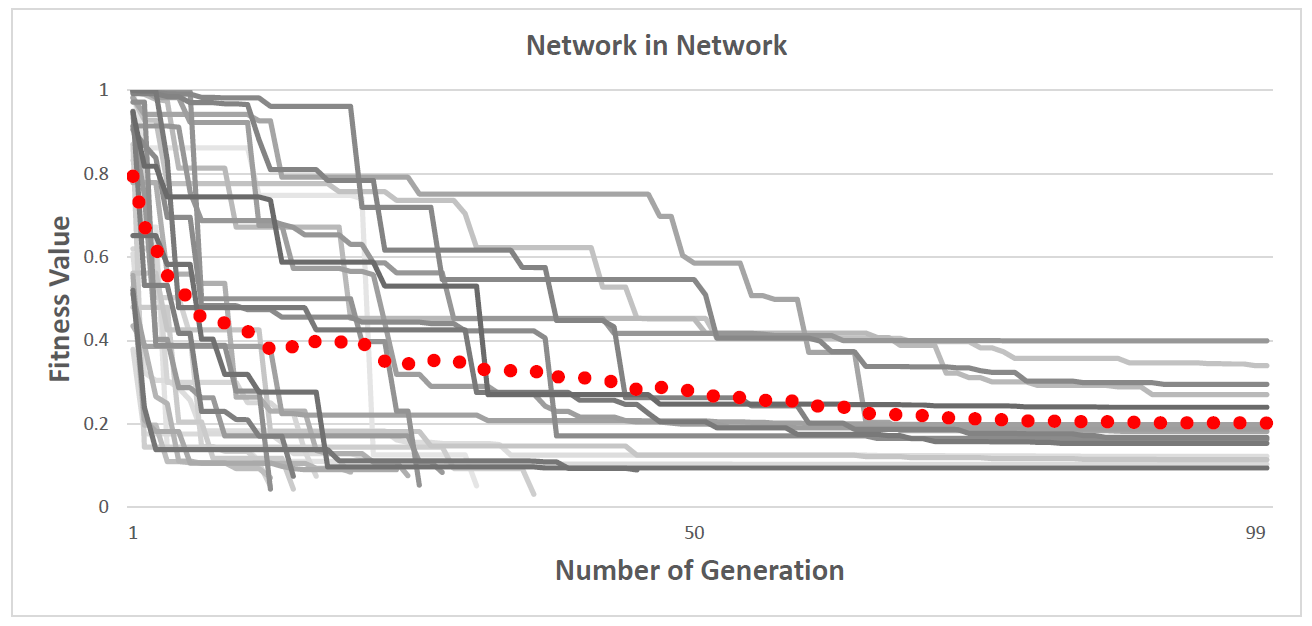}

\end{center}
\begin{center}
\includegraphics[width=1.0\linewidth]{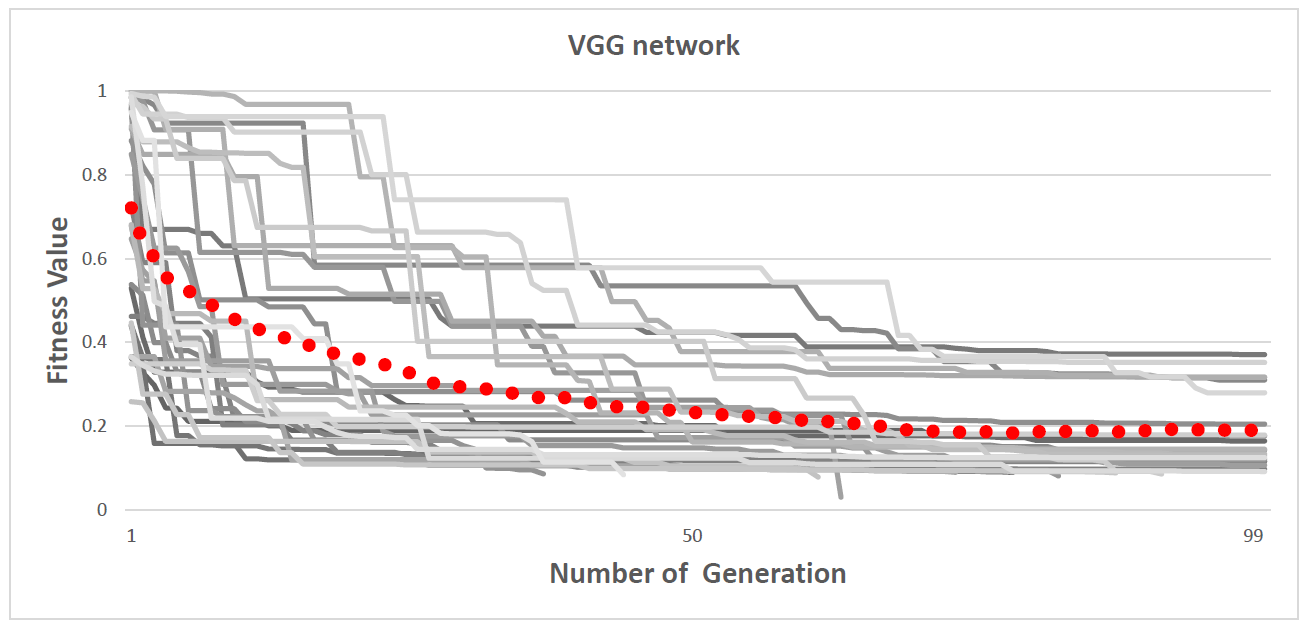}

\end{center}
\begin{center}
\includegraphics[width=1.0\linewidth]{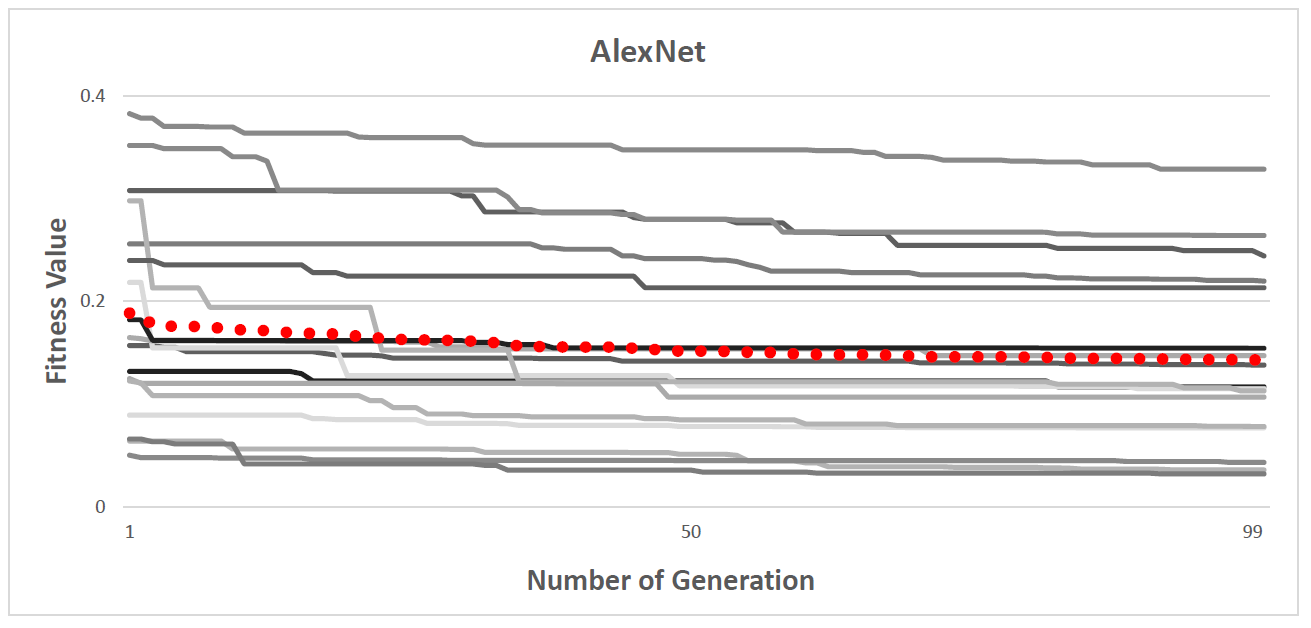}

\end{center}
   \caption{The change of fitness values during 100 generations of evolution of images (non-targeted) attacked by the proposed method among different network structures. The average values are highlighted by red dotted lines.}
\label{fig:long}
\label{fig:onecol} 
\end{figure}

\section{Results on Original CIFAR-10 Test Data}

We present another evaluation of one pixel attack which is on original CIFAR-10 test dataset \cite{11}. Comparing to the results on Kaggle CIFAR-10 aforementioned, the scenario is more limited since the images contain much less practical noise. Therefore, the target CNNs can have higher classification accuracy and confidence which definitely makes the attack harder.
Additionally, we only use images correctly classified by the target CNNs while in the experiment on Kaggle CIFAR-10 set we use all images (i.e., which contain wrongly classified images) with their true labels predicted by the target CNNs. 

%\textcolor{red}{Note that from the targeted attack, we find the phenomenon of semi-successful attack. That is, if targeted attacking an image with ground true class $A$ and target class $B$, the attack might end up with successfully perturbing the image to class $C$ where $A \neq B \neq C$. While this phenomenon is quite common, in our experiments on both kaggle and original CIFAR-10, we do not count it as a successful targeted attack. }

\begin{figure}[t]
\begin{center}
\includegraphics[width=0.8\linewidth]{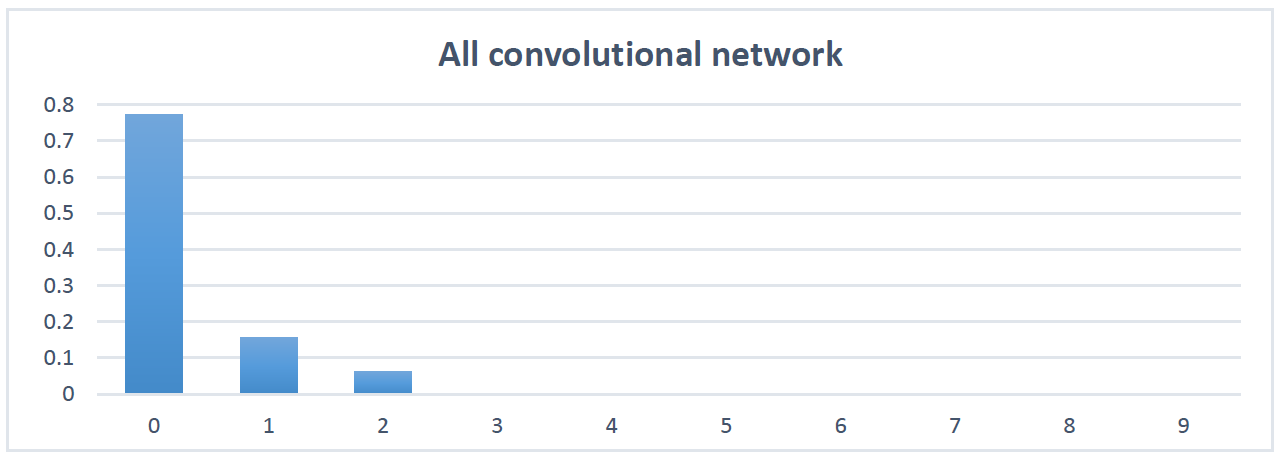}
\end{center}
\begin{center}
\includegraphics[width=0.8\linewidth]{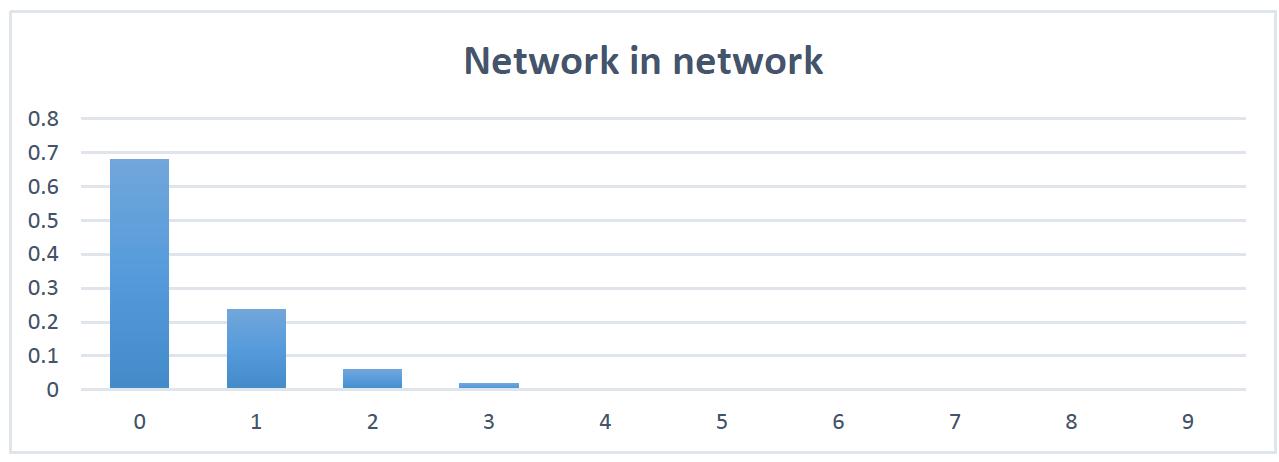}
\end{center}
\begin{center}
\includegraphics[width=0.8\linewidth]{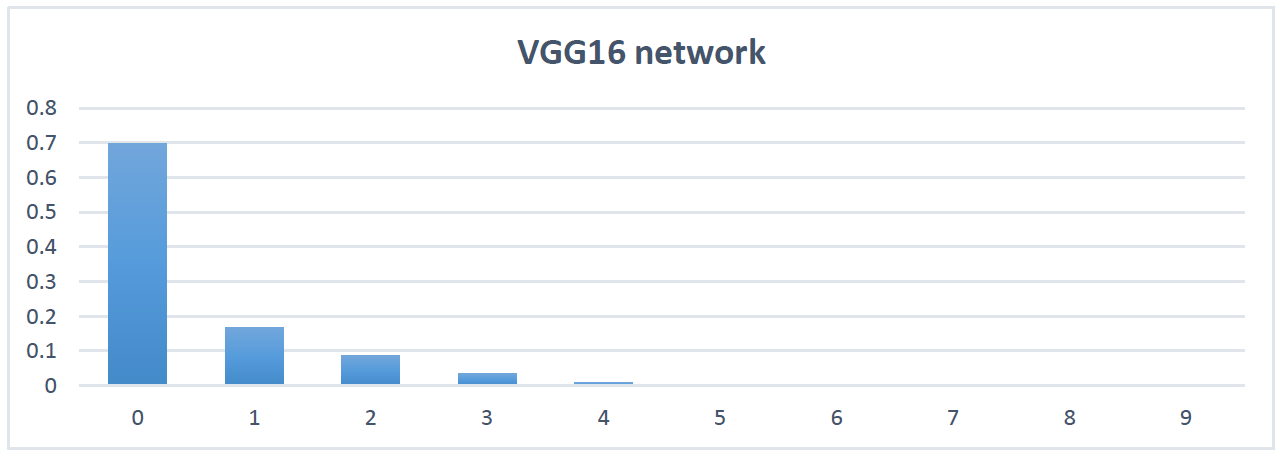}
\end{center}

   \caption{The percentage of natural images that were successfully perturbed to a certain number (from 0 to 9) of target classes by one pixel targeted attack.}

\label{fig:numclasses} 
\end{figure}

\begin{figure}[t]
\begin{center}
\includegraphics[width=0.6\linewidth]{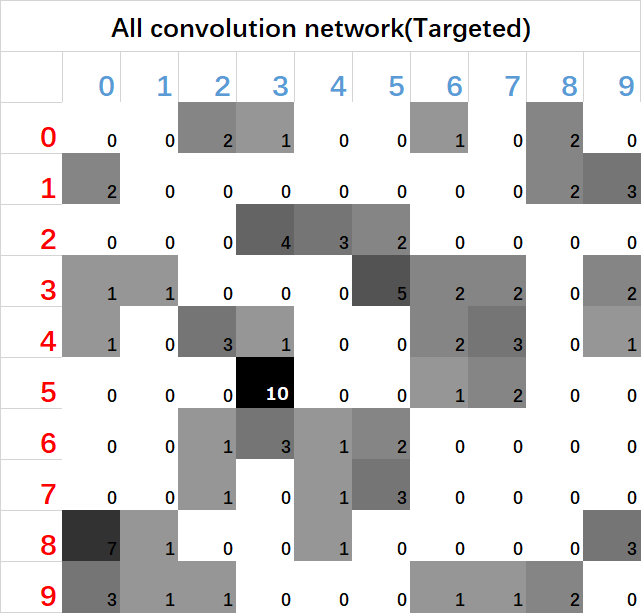}

\includegraphics[width=0.6\linewidth]{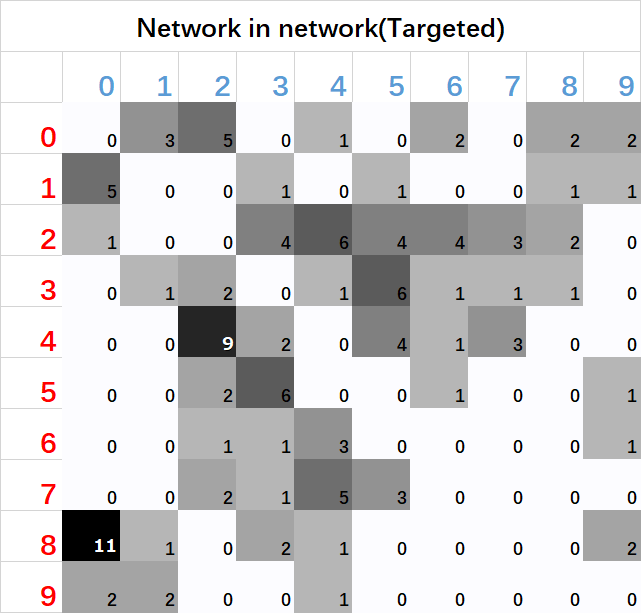}
\
\includegraphics[width=0.6\linewidth]{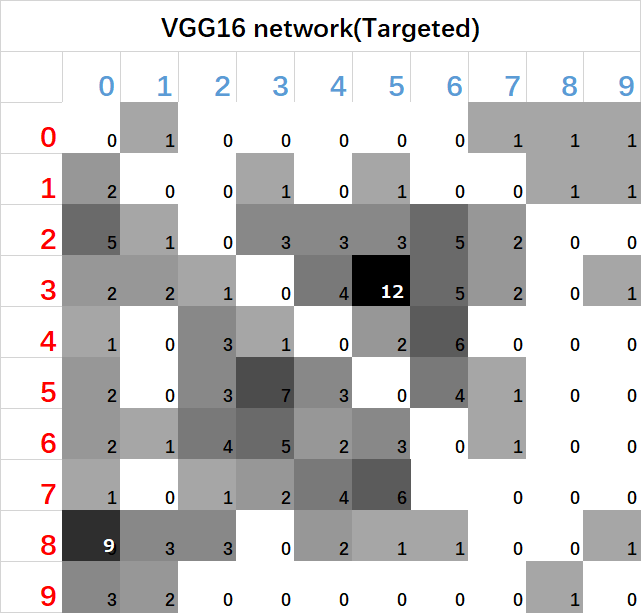}
\end{center}

   \caption{Heat-maps of the number of times a successful attack is present with the corresponding original-target class pair, for targeted attacks.}
\label{fig:heatmap}
\end{figure}

\begin{figure}[t]
\begin{center}
\includegraphics[width=0.6\linewidth]{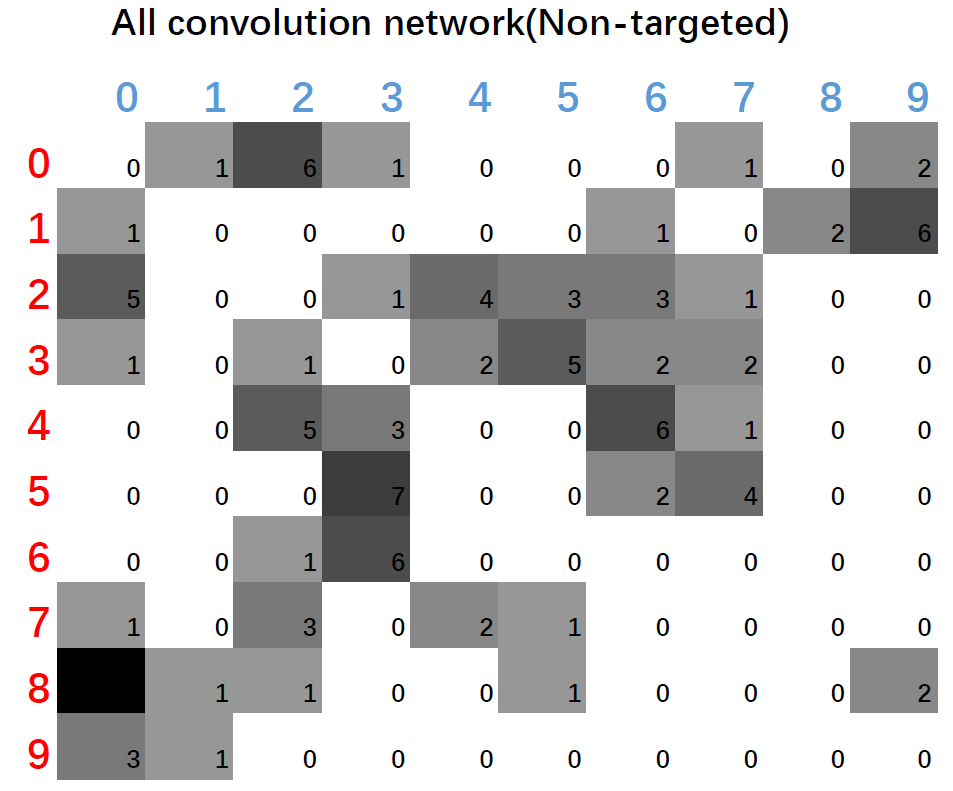}

\includegraphics[width=0.6\linewidth]{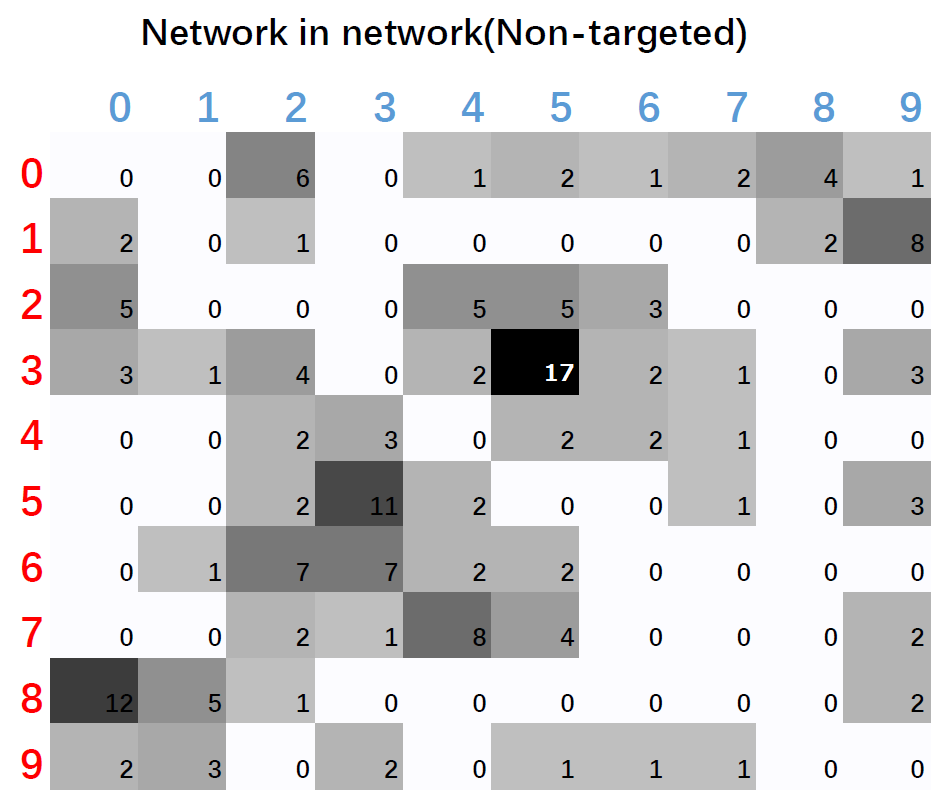}

\includegraphics[width=0.6\linewidth]{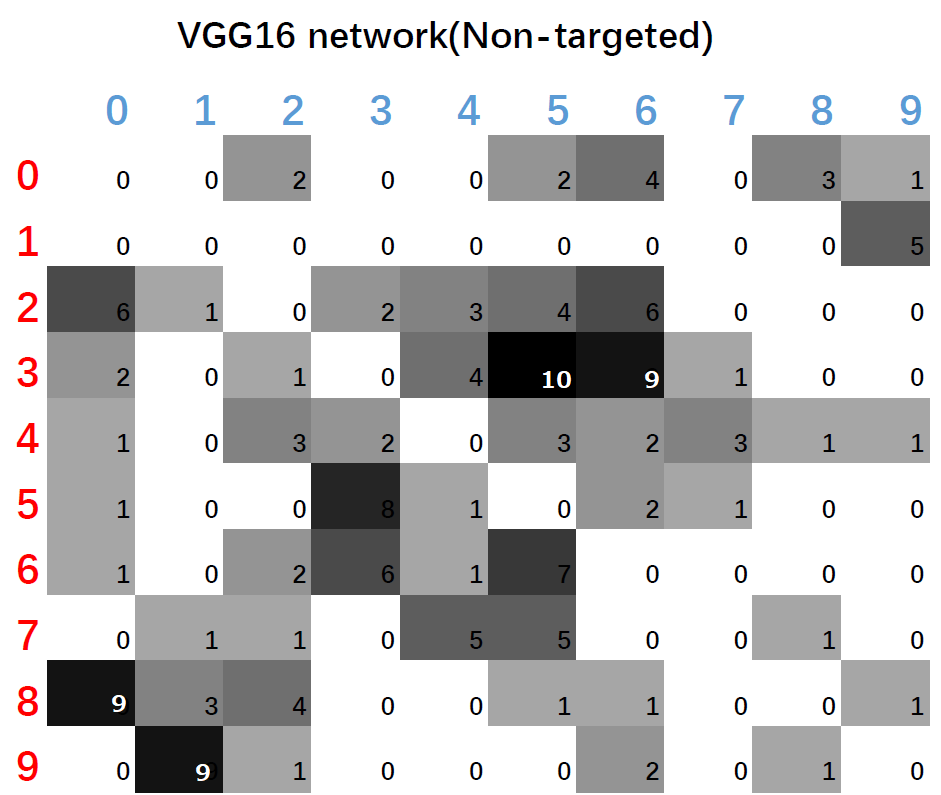}
\end{center}

   \caption{Heat-maps of the number of times a successful attack is present with the corresponding original-target class pair, for non-targeted attacks.}

\label{fig:heatmap1}
\end{figure}

\begin{figure}[t]
\begin{center}
\includegraphics[width=1.0\linewidth]{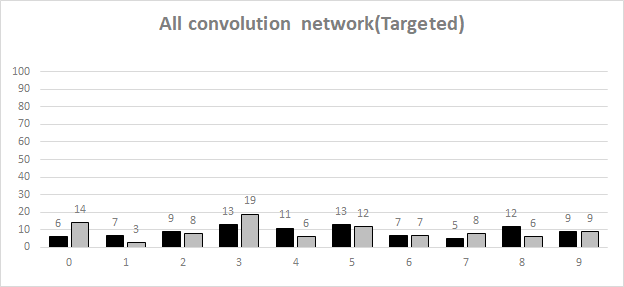}
\end{center}
\begin{center}
\includegraphics[width=1.0\linewidth]{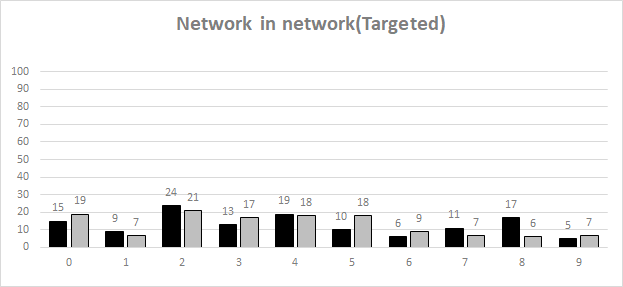}
\end{center}
\begin{center}
\includegraphics[width=1.0\linewidth]{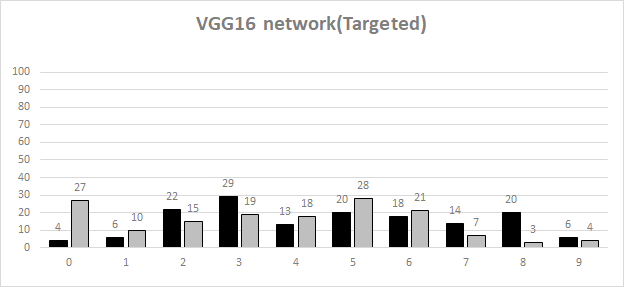}
\end{center}

   \caption{Number of successful attacks (vertical axis) for a specific class acting as the original (black) and target (gray) class, for targeted attacks.}

\label{fig:classes}
\end{figure}

\begin{figure}[t]
\begin{center}
\includegraphics[width=1.0\linewidth]{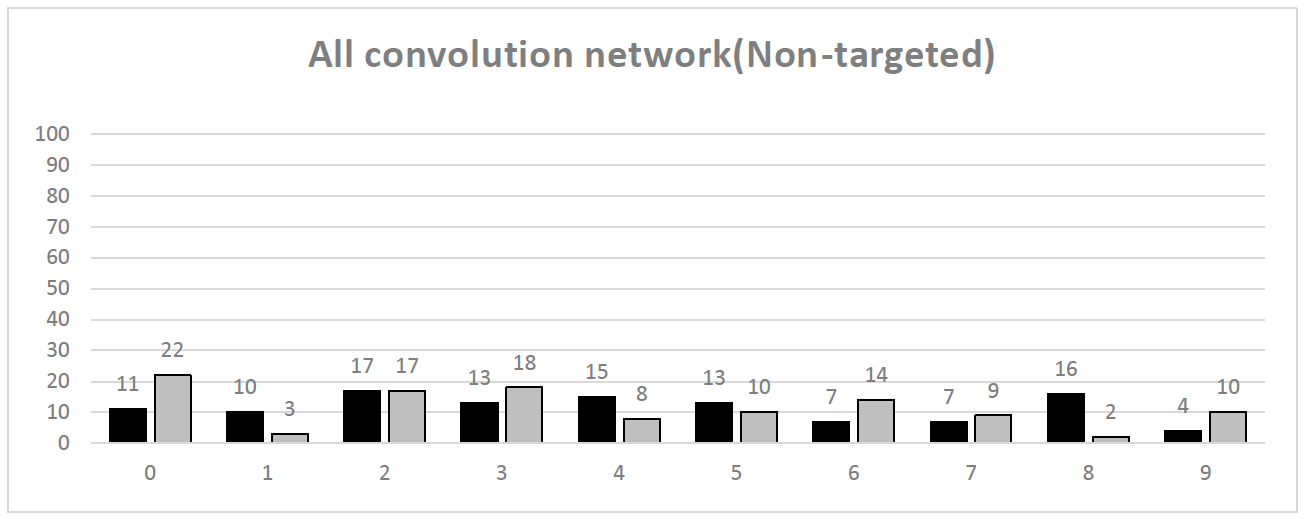}
\end{center}
\begin{center}
\includegraphics[width=1.0\linewidth]{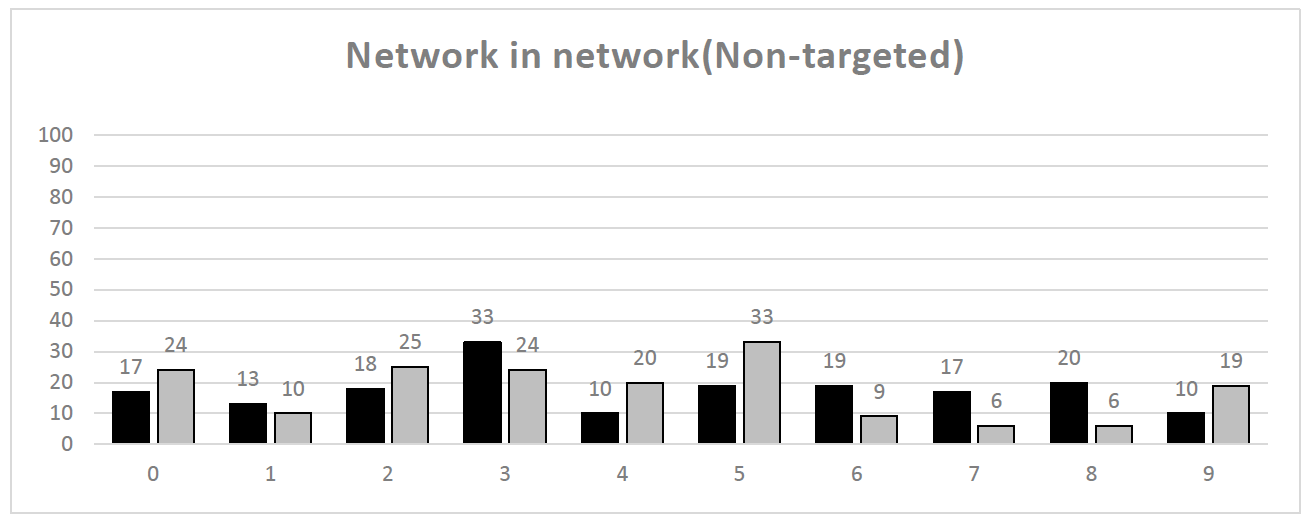}
\end{center}
\begin{center}
\includegraphics[width=1.0\linewidth]{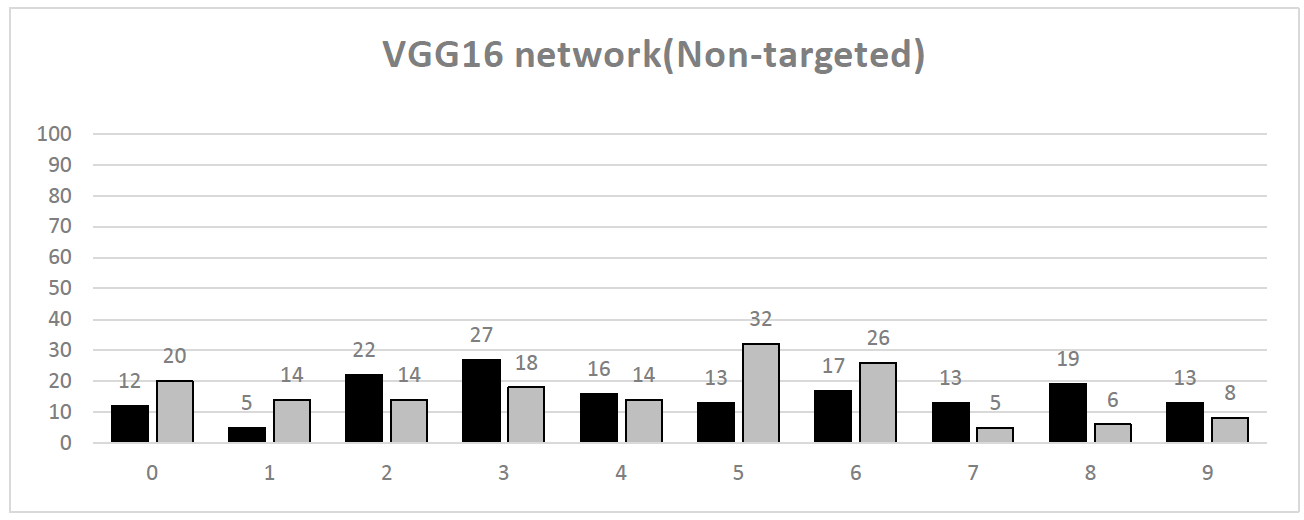}
\end{center}
   \caption{Number of successful attacks (vertical axis) for a specific class acting as the original (black) and target (gray) class, for non-targeted attacks.}

\label{fig:classes1}
\end{figure}

We use 500 random images for non-targeted attack and 300 for targeted attack. We also make small modification on network structure for better implementation. 
Specifically, for the Network in Network, we remove the second average pooling layers. For All convolutional network, we remove the batch normalization on the first layer. Three CIFAR-10 networks are re-trained to have similar natural accuracy to Table 4.
An early-stop criterion will be triggered when the probability label of the target class exceeds the original class.
All other settings are kept the same. 
The attack results are shown by Table 8. The number of target classes is shown by Figure~\ref{fig:numclasses}. 
The number of original-target class pairs is shown by the heat-maps of Figure~\ref{fig:heatmap} and Figure~\ref{fig:heatmap1}.
In addition to the number of original-target class pairs, the total number of times each class had an attack which either originated or targeted it is shown in Figure~\ref{fig:classes} and Figure~\ref{fig:classes1}.

\begin{table}
\begin{center}
\begin{tabular}{|c|c|c|c|c|}
	\hline  &AllConv&NiN&VGG16\\
	\hline Targeted&$3.41\%$&$4.78\%$&$5.63\%$\\
	\hline Non-targeted1&$22.67\%$&$32.00\%$&$30.33\%$\\
	\hline Confidence&$54.58\%$&$55.18\%$&$51.19\%$\\
	\hline Non-targeted2&$22.60\%$&$35.20\%$&$31.40\%$\\
	\hline Confidence&$56.57\%$&$60.08\%$&$53.58\%$\\
	\hline
\end{tabular}
\end{center}
\caption{Results of conducting one-pixel attack on original CIFAR-10 test set. Non-targeted1 indicates the non-targeted attack accuracy calculated from targeted attack results and Non-targeted2 indicates the true non-targeted attack accuracy. Other metrics are the same to Table 4.}
\end{table}

According to the attack results shown, we find the following features of one-pixel attack on original CIFAR-10. 
%We omit some explanation in detail since most of these features are similar to those have been found and analyzed on Kaggle CIFAR-10.}

1. Attack rate: The three networks have higher robustness to one-pixel attack according to the lower attack rate and confidence (Table~8). This might due to the higher classification accuracy and confidence of three networks on original CIFAR-10 test-set. Similar to the results on Kaggle set, the network in network still gets the lowest overall robustness considering both attack rate and confidence. 
This might be related to the proximity to the decision boundary. 
However, VGG network becomes much more vulnerable in this case. The discrepancy indicates that the robustness among different networks can be varied when handling images with low (e.g. Kaggle CIFAR-10) and high (e.g., original CIFAR-10) confidence. 

2.  Number of targeted classes: According to Figure~\ref{fig:numclasses}, it can be seen that in the case of targeted attack, it is still quite common that a vulnerable image can be perturbed to more than one class. In other words, the image might be locate near to the boundaries to multiple classes, especially in the case of VGG. This is similar to the Kaggle CIFAR-10 results shown by Fig~5. 

Note that one image can be perturbed to a final target class $A$ through the original target class $B$ (i.e. semi-successful targeted attack). For some images, the number of $B$ can be more than one.  We do not count it as a successful targeted attack unless $A$ = $B$.

3. Original-target class pairs: In both cases of targeted and non-targeted attack we again found the existence of vulnerable original-target classes pairs such as dog (5th)-cat (3rd) (Figure~\ref{fig:heatmap} and Figure~\ref{fig:heatmap1} ). In most cases, for a class pair between class $A$ and $B$, the number of successful perturbation from $A$ to $B$ is similar to the number of $B$ to $A$, which makes the heat-maps almost symmetric. However, there are exceptions such as ship (8th)-airplane (0th) pair, which the perturbation from ship to airplane class is very frequent but not vice versa.

Additionally, it also can be seen from Figure~\ref{fig:classes} and Figure~\ref{fig:classes1}, some vulnerable classes exist which have higher number of times being both original and target class of the attack.  A vulnerable original class is probably also vulnerable being a target class to a similar extend.

Most of these features, together with the specific vulnerable class-pairs shown by Figure~\ref{fig:heatmap} and Figure~\ref{fig:heatmap1} and vulnerable classes shown by Figure~\ref{fig:classes} and Figure~\ref{fig:classes1}, are similar or even exactly the same to the finding on attacking Kaggle CIFAR-10 dataset. 

\begin{table}
\begin{center}
\begin{tabular}{|c|c|c|c|c|}
	\hline Method&\rotatebox{90}{Success rate}&\rotatebox{90}{Confidence}&\rotatebox{90}{Number of pixels}&\rotatebox{90}{Network}\\

	\hline Our method&$35.20\%$&$60.08\%$&1 ($0.098\%$)&NiN\\
	\hline Our method&$31.40\%$&$53.58\%$&1 ($0.098\%$)&VGG\\
	\hline LSA\cite{13}&$97.89\%$&$72\%$&33 ($3.24\%$)&NiN\\
	\hline LSA\cite{13}&$97.98\%$&$77\%$&30 ($2.99\%$)&VGG\\
	\hline FGSM\cite{2}&$93.67\%$&$93\%$&1024 ($100\%$)&NiN\\
	\hline FGSM\cite{2}&$90.93\%$&$90\%$&1024 ($100\%$)&VGG\\

	\hline
\end{tabular}
\end{center}
\caption{Compassion of non-targeted attack effectiveness between the proposed method and two previous works. This suggests that one pixel is enough to create adversarial images from most of the natural images.}
\end{table}

\section{Discussion}

\subsection{Adversarial Perturbation}
%We disucuss some further correlations with related works.
Previous results have shown that many data points might be located near to the decision boundaries \cite{27}. For the analysis the data points were moved small steps in the input space while quantitatively analyzing the frequency of change in the class labels. In this paper, we showed that it is also possible to move the data points along few dimension to find points where the class labels change.
Our results also suggest that the assumption made by I. J. Goodfellow et al. that small addictive perturbation on the values of many dimensions will accumulate and cause huge change to the output \cite {2}, might not be necessary for explaining why natural images are sensitive to small perturbation. Since we only changed one pixel to successfully perturb a considerable number of images.

According to the experimental results, the vulnerability of CNN exploited by the proposed one pixel attack is generalized through different network structures as well as different image sizes.
In addition, the results shown here mimics an attacker and therefore uses a low number of DE iterations with a relatively small set of initial candidate solutions. Therefore, the perturbation success rates should improve further by having either more iterations or a bigger set of initial candidate solutions. Implementing more advanced algorithms such as Co-variance Matrix Adaptation Evolution Strategy \cite{99} instead of DE might also achieve the same improvement.
Additionally, the proposed algorithm and the widely vulnerable images (i.e. natural images that can be used to craft adversarial images to most of the other classes) collected might be useful for generating better artificial adversarial images in order to augment the training dataset. This aids the development of more robust models\cite{30} which is left as future work.

\subsection{Robustness of One-pixel Attack}

Some recently proposed detection methods have shown high accuracy of detecting adversarial perturbation. 
For example, B.Liang et al. utilize noise reduction to effectively detect both high and low-entropy images (e.g., bigger images give high entropy values) \cite{110}. In addition,  W. Xu et al. show that squeezing color bits and local/non-local spatial smoothing can simultanously detect $L_0$, $L_2$ and $L_\infty$ attacks \cite{109}. As the trade-off of being a low-cost, easy-implemented $L_0$ attack, we do not expect one pixel attack can achieve signficantly better robustness against such detection methods compared to other $L_0$ attacks such as \cite{99}.
%since we do not consider the corresponding countermeasures such as adding terms into fitness function of DE to favor resistance of these methods, as well as the fact that the resulting images of one -pixel attack are similar to other $L_0$ attacks that perturbs only few pixels. 

However, such detection schemes add another layer of pre-processing which increases the response time of the system. For example, both \cite{109} and \cite{110} require image processing and re-classification of the resulting images. Therefore they can be inefficient when dealing with adversarial scenarios such as novelty detection on security camera and image recognition systems on autonomous driving applications which run in real time with high frame rate. Besides, the impact of pre-processing on the classification accuracy is still not fully understood.

Detecting adversarial perturbation indeed can be helpful in practice. However, the fundamental problem is still left unsolved: the neural networks are still not able to recognize similar images as such, ignoring small adversarial perturbation. By proposing novel attack methods, we aim to emphasize the existence of different types of vulnerabilities and the corresponding understanding.
%more than the practical value of attacks themselves to which we believe that the much effort should also be paid.

\section{Future Work}

The DE utilized in this research belongs to a big class of algorithms called evolutionary strategies \cite{400} which includes other variants such as Adaptive DE \cite{401} and Covariance matrix adaptation evolution strategy (CMA-ES) \cite{402, 403, 404}. In fact, there are a couple of recent developments \cite{700, 701, 702} in evolutionary strategies and related areas that could further improve the current method, allowing for more efficient and accurate attacks.

%For example,  CMA-ES maintains and evolves a multivariate normal distribution to sample similar solutions in a single generation, whereas for one-pixel attack, it is reasonable to believe that a neighbor pixel with a similar RGB value to an existing attack pixel, can be also a solution, since both pixels can probably activate the same neurons to induce the misclassification. According to this property, we expect the search area can be large enough to ensure diversity of attacks during the early generations, then shrink the area to improve the quality of solution. CMA-ES can achieve this by its self-adaption mechanism of covariance matrix.

Furthermore, evolutionary computation also provides some promising approaches to solve adversarial machine learning related vulnerabilities. In fact, evolutionary-based machine learning allows for a great flexibility of models and may be an answer to the same problems it is revealing. First, in an area of evolutionary machine learning called neuroevolution, it was shown to be possible to learn not just the weights but also the topology of the network with evolutionary computation \cite{5, 244, 405}. In fact, SUNA \cite{5} goes beyond current neural models to propose a unified neuron model (e.g., time-scales, neuromodulation, feedback, long-term memory) that can adapt its structure and models to learn completely different problems (including non-markov problems) without changing any of its hyper-parameters. This generality is currently surpassing most if not all deep learning algorithms. Last but not least, self-organizing and novelty-organizing classifiers can adapt to changes in the environment by using flexible representations \cite{406, 407, 408}. For example, they can adapt to mazes that change in shape and to problems where the scope of variables change throughout the experiment \cite{409}: a very challenging scenario in which most if not all deep learning algorithms fail. These among other achievements \cite{410, 411} show a promising path that may solve current problems in deep neural networks in the years to come. 

Besides, it can be seen that the one-pixel attack can be potentially extended to other domains such as natural language processing, speech recognition, which will be also left for future work.

\section{Acknowledgment}
This research was partially supported by Collaboration Hubs for International Program (CHIRP) of SICORP, Japan Science and Technology Agency (JST), and Kyushu University Education and Research Center for Mathematical and Data Science Grant.

\ifCLASSOPTIONcaptionsoff
  \newpage
\fi

\clearpage

\bibliographystyle{IEEETrans}


\clearpage


%\vfill


\end{document}